\documentclass{article}
\PassOptionsToPackage{numbers,compress}{natbib}
\usepackage[dblblindworkshop, final]{neurips_2026}
\workshoptitle{Physical Understanding for Decision-Making: Bridging Foundation Models and Reliable Agents}

\usepackage[utf8]{inputenc} 
\usepackage[T1]{fontenc}    
\usepackage{hyperref}       
\usepackage{url}            
\usepackage{booktabs}       
\usepackage{amsfonts}       
\usepackage{nicefrac}       
\usepackage{microtype}      
\usepackage{xcolor}         
\usepackage{amsmath}
\usepackage{graphicx}
\usepackage{array}
\usepackage{wrapfig}
\usepackage{subcaption}
\usepackage[table]{xcolor}

\definecolor{rsqExcellent}{HTML}{1B7837}
\definecolor{rsqGood}     {HTML}{A6DBA0}
\definecolor{rsqOK}       {HTML}{FEE08B}
\definecolor{rsqWeak}     {HTML}{FFFFFF}
\definecolor{rsqBad}      {HTML}{FFFFFF}
\definecolor{rsqVeryBad}  {HTML}{FFFFFF}

\newcommand{\rsqA}[1]{\cellcolor{rsqExcellent}\textcolor{white}{\normalfont\textbf{#1}}}
\newcommand{\rsqB}[1]{\cellcolor{rsqGood}\textcolor{black}{\normalfont #1}}
\newcommand{\rsqC}[1]{\cellcolor{rsqOK}\textcolor{black}{\normalfont #1}}

\newcommand{\rsqD}[1]{\cellcolor{rsqBad}\textcolor{black}{\normalfont #1}}
\newcommand{\rsqE}[1]{\cellcolor{rsqBad}\textcolor{black}{\normalfont #1}}

\title{Evaluating Physical Consistency and Plausibility in Generative Scenario Models for Autonomous Driving}

\author{%
  Manasa Mariam Mammen$^{1,2}$ \And
  Zafer Kayatas$^{2}$ \And
  Stefan Wagner$^{1}$ \\[1em]
  \small
  $^{1}$Technical University of Munich, Heilbronn, Germany\\
  $^{2}$Mercedes Benz AG, Sindelfingen, Germany\\
}

\begin{document}

\maketitle

\begin{abstract}
Generative AI models are increasingly used for scenario generation in autonomous driving. While they can generate realistic looking scenarios, they often provide limited transparency into learned representations and consistency with real-world vehicle dynamics. This lack of formal assurance limits their use in safety-critical validation and certification workflows. To address this aspect, we introduce a layered evaluation protocol that complements existing methods by assessing models across five layers. The first four layers inspect internal representations and network layers through kinematic alignment, statistical baseline comparison, latent controllability, and activation analysis. The fifth layer evaluates model outputs against vehicle dynamics constraints such as lateral jerk thresholds. We demonstrate the protocol on a Variational Autoencoder (VAE) based scenario generator. Although standard output-level metrics and visualizations suggest that the generated scenarios are realistic, our protocol provides deeper insight into the extent to which the model’s latent space aligns with kinematic features and whether visually plausible trajectories satisfy vehicle-dynamics constraints. We further apply the protocol to additional generative models, demonstrating its applicability beyond the VAE architecture. 
\end{abstract}

\section{Introduction}
Recent progress in autonomous driving research, particularly in areas such as deep reinforcement learning and transfer learning, has substantially enhanced end-to-end perception and decision-making capabilities~\cite{hu2023planning,lee2022adas}. These advancements reflect the increasing complexity and adaptiveness of modern Advanced Driver Assistance System (ADAS) and autonomous systems, thereby increasing the need for more comprehensive and systematic validation methodologies~\cite{koopman2017autonomous}. 

Scenario-based testing in simulations plays a crucial role in validating the functionality and safety of Advanced Driving Systems (ADS)~\cite{zhong2021survey,menzel2018scenarios}. Instead of relying solely on road testing or statistical mileage accumulation, scenario-based approaches evaluate system behavior under structured and reproducible driving situations. These scenarios describe interactions between multiple traffic participants, environmental conditions, infrastructure layouts, and dynamic behaviors that may lead to safety-critical outcomes~\cite{gambi2022generating,ding2023survey,sauerbier2019definition}. Traditionally, scenario-based testing is resource-intensive and often relies on mathematical models that oversimplify real-world complexity, failing to capture the full spectrum of traffic scenarios. 

To address these limitations, the automotive industry is increasingly leveraging Artificial Intelligence (AI)~\cite{krajewski2019beziervae,chang2024llmscenario,lu2024realistic,xu2025diffscene,russell2025gaia} to enhance scenario-based testing.
AI methods offer the potential to model complex, real-world scenarios with greater accuracy by learning from large volumes of driving data. By integrating AI, it is possible to create more realistic and varied driving scenarios that better reflect the unpredictability and complexity of actual driving environments. Moreover, AI-powered scenario generation presents significant opportunities to explore rare and extreme events that are crucial for stress-testing ADAS and ensuring their reliability in unpredictable real-world conditions. 
 
Despite these opportunities, the industrial transferability of these models remains limited~\cite{gao2025survey}. A key reason is their lack of transparency. Unlike mathematical or rule-based scenario models, where experts can inspect assumptions, constraints, and generation logic directly, AI models often provide limited visibility into how scenarios are produced. Therefore, output-level realism checks alone are insufficient for validation. Test engineers also need evidence that the model has learned driving-relevant structures, can be reliably steered toward targeted scenario variations, and can produce trajectories that remain executable under vehicle dynamics constraints. Current evaluations mainly focus on reconstruction error, visual plausibility, or distributional similarity. However, these criteria do not sufficiently assess whether generated scenarios are structurally meaningful, controllable, or physically executable. As a result, scenarios that appear realistic at the output level may still be unsuitable for industrial testing. Addressing this gap requires interpretability methods. While tools such as probing~\cite{alain2016understanding}, latent traversal~\cite{fil2021beta}, and mechanistic interpretability~\cite{olah2020zoom,elhage2021mathematical} have been used to study representations in image and language models, their adaptation to generative scenario modeling remains limited.

In this paper, we introduce a structured evaluation protocol that directly targets these gaps for generative scenario models in autonomous driving. The protocol defines five complementary evaluation layers covering latent alignment, structure beyond statistical baselines, controllability, layer-wise maneuver organization, and physical executability We instantiate the protocol on a VAE trained on real-world highway maneuvers. The VAE's explicit latent space \(z\) offers the most direct and interpretable unit through which to explain the protocol's logic. However, the protocol itself is not specific to VAEs. It can be transferred to other generative architectures, as shown in the Appendix~\ref{app:additional_implementations} with appropriate adjustments to how representation layers are defined.

\section{Evaluation Setup}
\label{sec:setting}

The protocol requires three components : the generative model we want to evaluate, the real-world dataset it was trained, and a kinematic parameter extractor to probe for physically meaningful representations.

\paragraph{Model.}
The model used for the protocol demonstration is a VAE with a ten-dimensional latent space. The encoder maps a maneuver $\mathbf{x} \in \mathbb{R}^{100 \times 3}$ to an approximate posterior $q(\mathbf{z} \mid \mathbf{x})$, and the decoder reconstructs $\tilde{\mathbf{x}}$. We adopt a guided variant in which one latent dimension, $z_0$, is dedicated to scenario type and trained against a class-conditional Gaussian prior with an auxiliary alignment loss. The remaining dimensions $z_1$ through $z_9$ are free and follow a standard Gaussian prior. The training objective thus combines reconstruction error, the standard KL term, and the alignment loss:
\begin{equation}
    \mathcal{L}_{\text{total}} = \mathrm{MSE}(\tilde{\mathbf{x}}, \mathbf{x}) + \beta \, D_{\mathrm{KL}}\!\left(q(\mathbf{z}) \parallel p(\mathbf{z})\right) + \mathcal{L}_{\text{align}}.
\end{equation}
This design separates \emph{what to generate}, encoded in $z_0$, from \emph{how to generate it}, encoded in $z_1$ through $z_9$. The model exhibits strong performance across standard output-level evaluations, as reported in Appendix~\ref{app:model_details}, where further details on the architecture, training configuration, validation results, and initial latent space inspection are also provided.

\paragraph{Dataset.}
The model is trained on approximately 44{,}600 real world highway maneuvers collected on German highways, totalling about 1{,}735 hours of recording. Maneuvers are labeled into six safety-critical scenario types: cut-in left (CIL), cut-in right (CIR), cut-out left (COL), cut-out right (COR), cut-through left (CTL), and cut-through right (CTR). These scenario types describe how nearby vehicles move with respect to the ego vehicle. In a cut-in maneuver, a nearby vehicle changes lane into the ego vehicle’s lane. This may require the autonomous vehicle to react to a sudden intrusion. In a cut-out maneuver, a vehicle leaves the ego vehicle’s lane. This can create an unexpected gap or reveal a hidden traffic situation ahead, such as slow or stopped traffic. In a cut-through maneuver, a nearby vehicle crosses multiple lanes in one continuous motion. This scenario is more complex as it involves larger lateral movement and interactions across several lanes. More details about the dataset is given in Appendix~\ref{app:dataset_prep}

\paragraph{Physical Parameters.}
The physical parameters are extracted using the three-phase parametric driving model \cite{shamir2004should,salvucci2004inferring}. The full model description, including the equations and phase definitions, is provided in Appendix~\ref{app:mathematical_model}. The extracted parameter set serves as a representative baseline and can be updated, extended, or replaced according to the requirements of a specific user, dataset, or application.

\section{Proposed Evaluation Protocol}
\label{sec:methodology}

The protocol has five layers, each answering one question about the model. Layers 1-4 inspect internal representations and network layers. Layer~5 inspects generated scenarios against physical constraints.

\begin{enumerate}
    \item \textbf{Layer 1.} Does the latent space align with interpretable physical motion descriptors?
    \item \textbf{Layer 2.} Is the kinematic structure observed in Layer~1, a product of learning or a reflection of the data itself?
    \item \textbf{Layer 3.} How does the decoder transform the latent representation into a maneuver and where does scenario-level structure emerge?
    \item \textbf{Layer 4.} Can interventions on the latent representation reliably steer generated trajectories toward user-specified physical targets?
    \item \textbf{Layer 5.} Are the generated trajectories physically realizable by a real vehicle?
\end{enumerate}
\paragraph{Formal setup.}
Let \(g: \mathbb{R}^d \rightarrow \mathbb{R}^{T \times C}\) denote the decoder mapping a latent vector $\mathbf{z}$ to a maneuver of $T$ timesteps and $C$ channels. In this work, $d = 10$, $T = 100$, and $C = 3$ (time, lateral position, longitudinal velocity). For a dataset of $N$ maneuvers, $\{\mathbf{z}^{(i)}\}_{i=1}^{N}$ are the latent representations and $\{\mathbf{p}^{(i)}\}_{i=1}^{N}$ the corresponding physical parameters extracted with $\mathbf{p}^{(i)} \in \mathbb{R}^M$ containing $M = 12$ physical quantities (9 independent and 3 derived).

\subsection{Layer 1: Latent-Physical Probing}
\label{sec:probing}

Probing trains a simple regressor on a frozen representation to measure what information it encodes~\cite{alain2016understanding}. We adapt it to physical regression: for each parameter $p_j$ in $\mathbf{p}$, a probe $f_j: \mathbb{R}^d \rightarrow \mathbb{R}$ predicts $p_j$ from $\mathbf{z}$, evaluated by $R^2$ under five fold cross validation.

Two probes of different complexity are used. A linear probe fits a Ridge regression~\cite{hoerl1970ridge}, $\hat{p}_j = \mathbf{w}^\top \mathbf{z} + b$, which tests whether $p_j$ is encoded as a linear function of $\mathbf{z}$. A nonlinear probe fits a one hidden layer MLP, $\hat{p}_j = \mathbf{v}^\top \phi(\mathbf{W} \mathbf{z} + \mathbf{b}_1) + b_2$ with ReLU activation, which tests whether $p_j$ is encoded in a nonlinear or distributed form. The probing gap $\Delta R^2 = R^2_\text{nonlinear} - R^2_\text{linear}$ quantifies nonlinear entanglement: a small gap indicates that linear probes recover what is present, a large gap indicates encoded information is entangled across dimensions.

To localise where in the latent space each parameter is encoded, we additionally fit per dimension univariate regressions, $\hat{p}_j = w_k z_k + b_k$, yielding a $d \times M$ matrix of $R^2$ values. This matrix distinguishes disentangled encoding (a parameter localized to one dimension) from entangled encoding (a parameter distributed across dimensions). As a null model, the same procedure is applied to random vectors drawn from a standard normal, confirming that high $R^2$ values reflect genuine representations rather than artifacts of the probing procedure.

\subsection{Layer 2: Baseline Comparison}
\label{sec:pca_baseline}

A high probing $R^2$ alone does not show that the model learned anything meaningful, since any compression that preserves the data well could yield recoverable parameters. We compare against Principal Component Analysis (PCA)~\cite{mackiewicz1993principal}, which provides an optimal linear compression of the maneuver data in the mean squared error sense and captures all variance structure that purely statistical compression can extract.

Each maneuver $\mathbf{x}^{(i)} \in \mathbb{R}^{T \times C}$ is flattened and projected onto the first $d$ principal components to obtain score vectors $\mathbf{s}^{(i)} \in \mathbb{R}^d$. The probing procedure of Layer~1 is applied to $\mathbf{s}^{(i)}$, and the model's advantage on parameter $p_j$ is defined as $A_j = R^2_{\text{model},j} - R^2_{\text{PCA},j}$. A positive $A_j$ indicates that the learned representation is more aligned with $p_j$ than purely statistical compression achieves and  a near-zero advantage means the model offers no interpretability benefit over PCA for that parameter.

\subsection{Layer 3: Decoder Analysis}
\label{sec:decoder}
Layers 1 and 2 characterize the features learned by the model and assess their interpretability. Layer 3 now opens the decoder to study how these features are transformed into a driving maneuver by its internal mechanisms. Following mechanistic interpretability principles~\cite{olah2020zoom}, we trace how the latent representation is progressively transformed across the decoder's hidden layers using two complementary analyses: 
\begin{itemize}
    \item Layer-wise activation clustering which reveals where scenario-level structure emerges.
    \item Layer-wise probing which measures how decodable physical parameters and the full trajectory are from each layer's representation.
\end{itemize} 

\paragraph{Activation clustering.}
For each decoder layer $l$, activations $\mathbf{h}^{(l)} \in \mathbb{R}^{n_l}$ are collected across $N$ samples spanning all six scenario types, forming an activation matrix $\mathbf{H}^{(l)} \in \mathbb{R}^{N \times n_l}$. The matrix is centered and projected to two dimensions via PCA~\cite{mackiewicz1993principal},
\begin{equation}
    \mathbf{Z}^{(l)} = \left(\mathbf{H}^{(l)} - \bar{\mathbf{h}}^{(l)}\right) \mathbf{U}_2^{(l)}, \qquad \mathbf{U}_2^{(l)} \in \mathbb{R}^{n_l \times 2},
    \label{eq:pca_projection}
\end{equation}
where $\bar{\mathbf{h}}^{(l)}$ is the mean activation at layer $l$ and $\mathbf{U}_2^{(l)}$ contains the two leading eigenvectors of the layer's empirical covariance matrix. The two dimensional projection $\mathbf{Z}^{(l)} \in \mathbb{R}^{N \times 2}$ is used for visualisation and for cluster quality assessment.

\paragraph{Layer wise probing.}
At each decoder layer $l$, two linear probes are fit on $\mathbf{h}^{(l)}$. A parameter probe predicts each scalar physical quantity $p_j$, $\hat{p}_j^{(l)} = \mathbf{w}_j^{(l)\top} \mathbf{h}^{(l)} + b_j^{(l)}$. A trajectory probe predicts the full output, $\hat{\mathbf{y}}^{(l)} = \mathbf{W}^{(l)} \mathbf{h}^{(l)} + \mathbf{b}^{(l)}$. Both are evaluated by $R^2$ on a held out split. Together they ask complementary questions: trajectory probing measures \emph{how much} of the output is linearly recoverable at each layer, while parameter probing measures \emph{which} physical quantities are explicitly at each layer revealing whether a quantity is encoded from the outset or built up progressively through the network (The layer at which a parameter first achieves high $R^2$ serves as a localization signal, indicating the depth at which that physical quantity emerges in the model's internal representations or pointing to specific circuits in transformer-based models). If both achieve high $R^2$ from early layers onward, the decoder does not construct maneuvers through progressive nonlinear composition but rather encodes physical parameters linearly from the outset, with the trajectory reducible to their linear combination.

\subsection{Layer 4: Controllability via Latent Intervention}
\label{sec:sensitivity}

Probing alone does not establish that the decoder \emph{uses} the encoded information: a high $R^2$ can coexist with information the decoder ignores. Layer~4 therefore intervenes on the latent space and observes the resulting change in physical parameters. We combine three complementary analyses: latent traversal to test directional consistency, polynomial fitting to characterise the form of the response, and nudge based sensitivity to produce practical control guidance.

\paragraph{Latent traversal.}
Following the traversal protocol from disentanglement analysis~\cite{burgess2018understanding}, we vary one latent dimension at a time around the reference point $\mathbf{z}_0$ (the training mean):
\begin{equation}
    \mathbf{z}(\alpha) = \mathbf{z}_0 + \alpha \, \sigma_k \, \mathbf{e}_k, \qquad \alpha \in [-3, 3],
    \label{eq:traversal}
\end{equation}
where $\sigma_k$ is the standard deviation of $z_k$ across the training set. The traversal is decoded, the physical parameters are extracted, and directional consistency is measured by the Pearson correlation between traversal magnitude and parameter change, $\rho_{k,j} = \mathrm{corr}(\alpha, p_j(\alpha))$. A high $|\rho_{k,j}|$ indicates that the decoder responds to $z_k$ in a coherent and directionally predictable way.

\paragraph{Functional form.}
Directional consistency does not characterise whether the response is proportional or curved. To capture the form, we fit polynomials of increasing degree to the traversal response,
\begin{equation}
    \Delta p_j(\delta) \approx \sum_{n=1}^{D} c_n \, \delta^n,
    \label{eq:poly_fit}
\end{equation}
sweeping $D \in \{1, 2, 3\}$ and selecting the lowest $D^\star$ that achieves $R^2 > 0.95$, requiring at least a $0.5\%$ improvement over the next lower degree. $D^\star = 1$ corresponds to proportional control, where doubling a perturbation doubles the response and steering can be planned with simple linear rules. $D^\star \geq 2$ indicates a curved transfer function in which the response is superlinear or sublinear in the perturbation magnitude, and reliable steering requires either a calibrated nonlinear inverse map or operating point dependent gain compensation.

\paragraph{Practical sensitivity.}
The traversal and polynomial analyses describe the form of the response. The third analysis quantifies its magnitude in engineering terms. For each dimension $z_k$ and nudge magnitude $\delta \in \{-0.5, -0.2, -0.1, 0.1, 0.2, 0.5\}$, we decode both $\mathbf{z}_0$ and $\mathbf{z}_\delta = \mathbf{z}_0 + \delta \mathbf{e}_k$ and report the change in physical parameter $p_j$ normalised by its population standard deviation,
\begin{equation}
    \Delta \tilde{p}_j(\delta, k) = \frac{p_j(\mathbf{z}_\delta) - p_j(\mathbf{z}_0)}{\sigma_{p_j}}.
    \label{eq:normalized_change}
\end{equation}
A change is practically significant when $|\Delta \tilde{p}_j| > \tau$ with $\tau = 0.1$, motivated by effect size conventions in statistical analysis~\cite{cohen1988statistical}. The triple of $(\rho_{k,j}, D^\star, \Delta \tilde{p}_j)$ jointly characterises a candidate control axis: $\rho_{k,j}$ tells whether the response is directionally consistent, $D^\star$ tells whether it is linear or curved, and $\Delta \tilde{p}_j$ tells how much physical change a given nudge produces. Together these provide a per dimension control guide indicating which latent dimension an engineer should adjust for a target physical change, whether that change can be steered with simple linear rules or requires nonlinear correction, and how large a perturbation is needed to achieve a practically significant effect.

\subsection{Layer 5: Physical Plausibility Evaluation}
\label{sec:physical_plausibility}

The first four layers characterize how the latent space encodes and controls maneuver parameters. Layer~5 asks a complementary question: do the trajectories the decoder produces obey the physical constraints of a real vehicle? Existing benchmarks rely on displacement based metrics which measure closeness to a reference distribution rather than executability under tire friction, actuator, and stability limits. Unless these constraints are explicitly used during training, they are difficult for the model to learn from data alone. We therefore want to test whether the decoder has implicitly internalized such constraints, or whether it can generate trajectories that appear realistic in displacement space but are physically infeasible.

From the generated scenario trajectories, we compute longitudinal and lateral accelerations and lateral jerk by central differences, and check three thresholds: a friction circle constraint $\sqrt{a_x^2 + a_y^2}/g \leq \mu$ with $\mu = 0.8$ for dry asphalt~\cite{pacejka2005tire}; a lateral acceleration limit $\max_t |a_y| \leq 3\,\text{m/s}^2$ for comfortable highway maneuvering and a lateral jerk limit $\max_t |j_y| \leq 5.0\,\text{m/s}^3$ for kinematic feasibility as mentioned in ISO 11270.

The Physical Validity Rate (PVR) for scenario $s$ is the fraction of trajectories satisfying all three constraints. Because the training data itself contains aggressive maneuvers exceeding comfort thresholds, a perfect model is not expected to reach $\mathrm{PVR} = 100\%$. We therefore report the relative rate
\begin{equation}
    \Delta_s = \mathrm{PVR}_s^\text{generated} - \mathrm{PVR}_s^\text{train},
    \label{eq:pvr_delta}
\end{equation}
which distinguishes generation artifacts from distribution faithful reproduction of the aggressive driving tail. A near zero $\Delta_s$ indicates faithful reproduction; strongly negative values expose a learning failure invisible to displacement based metrics; strongly positive values indicate the model is more conservative than the data. 

\section{Evaluation Results on Guided Variational Autoencoder}
\label{sec:results}

We now apply the proposed protocol to the guided VAE and assess its interpretability, controllability, and physical feasibility (generated scenarios).. To keep the discussion focused, analyses that add only limited additional information beyond the primary findings for a given layer are included in the Appendix~\ref{app:further_layers}.

\subsection{Layer 1: Latent-Physical Probing}
\label{sec:results_l1}
\begin{wraptable}{r}{0.5\textwidth}
    \scriptsize
    \setlength{\tabcolsep}{4pt}
    \caption{Probing $R^2$ for physical parameters from $\mathbf{z}$. Bold: $R^2_\text{linear} \geq 0.7$.}
    \label{tab:probing}
    \begin{tabular}{l c c c c c} \hline \textbf{Parameter} & $R^2_\text{lin}$ & $R^2_\text{nl}$ & $\Delta R^2$ & $R^2_\text{null,lin}$ & $R^2_\text{null,nl}$ \\ \hline $y_\text{init}$ & \textbf{0.952} & 0.989 & 0.038 & -0.001 & -0.003 \\ $y_\text{final}$ & \textbf{0.957} & 0.992 & 0.035 & -0.001 & -0.004 \\ $v_\text{start}$ & \textbf{0.941} & 0.977 & 0.036 & 0.000 & -0.007 \\ $v_\text{end}$ & \textbf{0.951} & 0.977 & 0.027 & 0.000 & -0.005 \\ $\Delta v$ & \textbf{0.756} & 0.888 & 0.132 & -0.001 & -0.001 \\ $\bar{\dot{y}}$ & \textbf{0.811} & 0.906 & 0.095 & -0.001 & -0.012 \\ $a_\text{pre}$ & 0.195 & 0.631 & 0.435 & 0.000 & -0.005 \\ $a_\text{dur}$ & 0.538 & 0.665 & 0.126 & -0.001 & -0.016 \\ $a_\text{post}$ & 0.235 & 0.498 & 0.263 & -0.001 & -0.011 \\ $t_\text{pre}$ & 0.097 & 0.502 & 0.405 & -0.001 & -0.009 \\ $t_\text{dur}$ & 0.071 & 0.456 & 0.385 & -0.001 & -0.008 \\ $t_\text{post}$ & 0.111 & 0.658 & 0.548 & -0.001 & -0.009 \\ \hline \end{tabular}
\end{wraptable}
Table~\ref{tab:probing} reports cross-validated $R^2$ for predicting twelve physical parameters from the latent vector $\mathbf{z}$, together with the probing gap $\Delta R^2$ and the null model values.
The parameters are split into three categories. Maneuver boundary conditions ($y_\text{init}$, $y_\text{final}$, $v_\text{start}$, $v_\text{end}$) are encoded almost perfectly linearly with $R^2_\text{linear} \geq 0.94$ and $\Delta R^2 < 0.04$, meaning they are directly readable from $\mathbf{z}$ as linear combinations of its dimensions. Aggregate derived quantities ($\Delta v$, $\bar{\dot{y}}$) remain linearly accessible but with mild nonlinear entanglement. In case of phase-wise accelerations and durations behave very differently: linear probes are weak (durations near chance, $R^2 \leq 0.11$), while nonlinear probes recover substantial variance (gaps up to $0.55$). The model encodes \emph{when} events occur, but not in any linearly decodable form. We
interpret this asymmetry as evidence that the latent space encodes the
\emph{continuous} driving dynamics directly, rather than the
piecewise-linear discretization imposed by the three-phase
parameterization. The null model $R^2$ stays between $-0.016$ and $0.000$, consistent with cross-validated random features and confirming that the probing procedure does not inflate scores. The per-dimension probing matrix further shows that some of the linearly encoded parameters concentrate in one dominant dimension, but not exclusively. This suggests partially localized yet distributed representations of physical quantities as provided in more detail in Appendix~\ref{app:further_layer_1}.

\subsection{Layer 2: Baseline Comparison}
\label{sec:results_l2}
\begin{wraptable}{r}{0.5\textwidth}
    \centering
    \scriptsize
    \caption{Best single dimension probing $R^2$: VAE vs.\ PCA, with advantage $A_j$.}
    \label{tab:pca}
    \begin{tabular}{l c c c}
        \hline
        \textbf{Parameter} & $R^2_\text{VAE}$ & $R^2_\text{PCA}$ & $A_j$ \\
        \hline
        $y_\text{init}$   & 0.762 & 0.676 & $+0.086$ \\
        $y_\text{final}$  & 0.637 & 0.524 & $+0.113$ \\
        $v_\text{start}$  & 0.760 & 0.711 & $+0.049$ \\
        $v_\text{end}$    & 0.743 & 0.721 & $+0.022$ \\
        $\Delta v$        & 0.501 & 0.767 & $-0.266$ \\
        $a_\text{dur}$    & 0.327 & 0.520 & $-0.193$ \\
        $\bar{\dot{y}}$   & 0.587 & 0.647 & $-0.060$ \\
        \multicolumn{4}{c}{\textit{remaining params: $R^2 < 0.16$ for both}} \\
        \hline
    \end{tabular}
\end{wraptable}

Table~\ref{tab:pca} compares the best single-dimension probing $R^2$ of the VAE latent space with PCA on the raw maneuver data, using $A_j = R^2_\text{VAE} - R^2_\text{PCA}$. The VAE outperforms PCA on the four boundary conditions, especially lateral position ($A_j = +0.086$ for $y_\text{init}$, $+0.113$ for $y_\text{final}$). While PCA spreads lateral information across many components, the VAE reorganizes it into compact, axis-aligned coordinates, aided by reconstruction and the explicit conditioning of $z_0$ on scenario type. Velocity boundary conditions show smaller gains: although velocity dominates trajectory variance and is therefore well captured by PCA, the VAE represents both endpoints more compactly on a shared dimension, $z_5$.

For derived velocity quantities, PCA performs better, most notably for $\Delta v$ ($A_j = -0.266$) and $a_\text{dur}$ ($A_j = -0.193$). Since these quantities are linear functionals of the trajectory, they are naturally recovered by an orthogonal linear decomposition. The VAE instead places both on the shared axis $z_4$, reflecting their physical coupling but reducing single-parameter fidelity. The remaining parameters, including pre- and post-accelerations and all phase durations, are poorly recovered by both representations.

\subsection{Layer 3: Decoder Analysis}
\label{sec:results_l3}
The Layer 3 analysis examines how physical structure emerges within the decoder and how this structure relates to the final generated maneuver.
\paragraph{Activation clustering.}

\begin{figure}[h]
    \centering
    \includegraphics[width=0.65\linewidth]{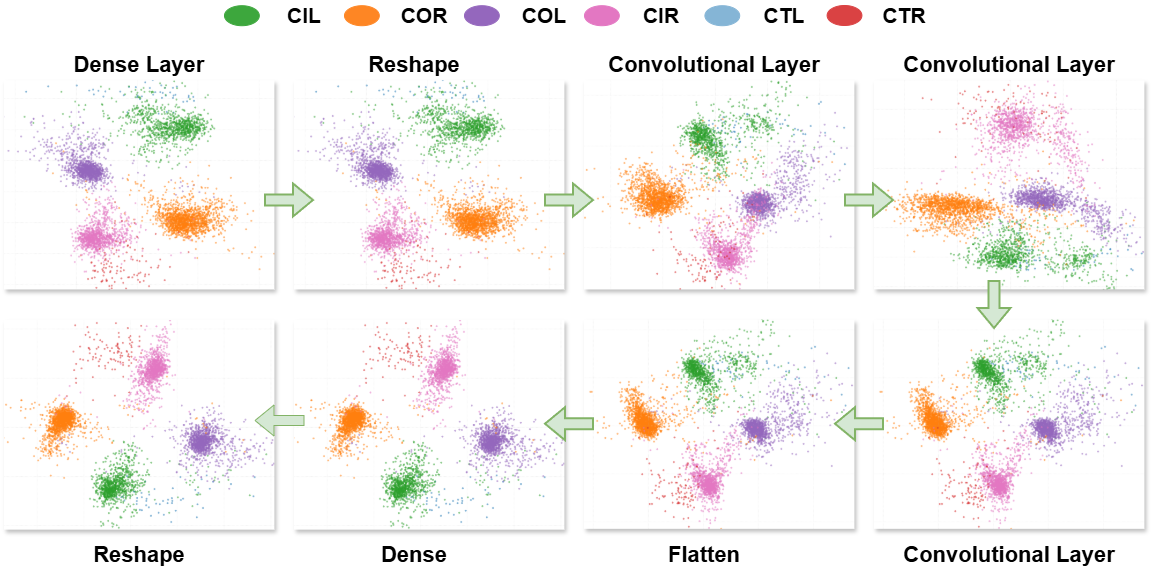}
    \caption{Visualization of cluster separation from early to late layers.}
    \label{fig:pca_layers}
\end{figure}

PCA projections of decoder activations (Figure~\ref{fig:pca_layers}) show that the six scenario types form distinct clusters from the first dense layer onward, and the discriminative structure is preserved even in the deep layers. A cut-through maneuver which is mechanically a composition of two single lane changes (CTL = CIL + COL; CTR = CIR + COR), at every layer the CTL cluster sits between the CIL and COL clusters in projection. These clusters suggest that each scenario produces a distinct activation pattern, which is confirmed by a unit level analysis of the first dense layer where the six considered scenario shows an almost reproducible activation signature, while left and right variants exhibit mirrored patterns (Appendix~\ref{app:further_layers}).
\paragraph{Layer-wise probing.}
Linear probes from decoder activations to physical parameters reveal a U-shaped pattern across depth (Table~\ref{tab:layer_param_probe}). Boundary parameters are highly recoverable at the latent input and first dense layer ($R^2 > 0.95$), with the dense layer slightly increasing the linear readability already present in $\mathbf{z}$. In the middle convolutional layers, probing performance collapses. Since the same parameters are again recovered with $R^2>0.99$ after the final dense layer, the information is preserved but represented in a form inaccessible to linear probes. After flattening and the final dense+tanh+reshape step, $R^2$ returns to values comparable to or above the first dense layer, indicating an information-preserving reorganization into engineering-aligned coordinates. The activation density $\rho^{(l)}$ follows the same U-shape, dropping from $0.30$ at the first dense layer to $0.04$ in the middle convolutional layers before rising to $1.0$ at the output. This supports the view that the decoder's convolutional middle performs sparse spatio-temporal computation in a representation not directly aligned with human-interpretable parameters.

\begin{table*}[h]
\centering
\caption{Layer-wise Parameter Probing.}
\label{tab:layer_param_probe}

\setlength{\tabcolsep}{4pt}
\renewcommand{\arraystretch}{1.15}
\footnotesize

\begin{tabular}{lccccccccc}
\toprule
\textbf{Parameter}
& {\shortstack{\textbf{z}}}   
& {\shortstack{Dense}}
& {\shortstack{Reshape}}
& {\shortstack{Convolutional}}
& {\shortstack{Convolutional}}
& {\shortstack{Convolutional}}
& {\shortstack{Flatten}}
& {\shortstack{Dense}}
& {\shortstack{Reshape}}
\\
\midrule
 Activation density & 100\% & 30\% & 30\% & 10\% & 4\% & 4\% & 
                 20\% & 100\% & 100\% \\
y\_init          & \rsqA{+0.952} & \rsqA{+0.989} & \rsqA{+0.989} & \rsqA{+0.970} & \rsqD{+0.102} & \rsqA{+0.985} & \rsqA{+0.985} & \rsqA{+0.991} & \rsqA{+0.991} \\
y\_final         & \rsqA{+0.957} & \rsqA{+0.992} & \rsqA{+0.992} & \rsqA{+0.982} & \rsqB{+0.930} & \rsqA{+0.990} & \rsqA{+0.990} & \rsqA{+0.994} & \rsqA{+0.994} \\
v\_start         & \rsqB{+0.941} & \rsqA{+0.979} & \rsqA{+0.979} & \rsqB{+0.935} & \rsqE{<0}& \rsqA{+0.969} & \rsqA{+0.969} & \rsqA{+0.982} & \rsqA{+0.982} \\
v\_end           & \rsqA{+0.951} & \rsqA{+0.978} & \rsqA{+0.978} & \rsqB{+0.950} & \rsqE{<0}& \rsqB{+0.942} & \rsqB{+0.942} & \rsqA{+0.981} & \rsqA{+0.981} \\
delta\_v         & \rsqC{+0.756} & \rsqB{+0.889} & \rsqB{+0.889} & \rsqC{+0.769} & \rsqE{<0}& \rsqC{+0.744} & \rsqC{+0.744} & \rsqB{+0.907} & \rsqB{+0.907} \\
a\_pre           & \rsqD{+0.195} & \rsqC{+0.620} & \rsqC{+0.620} & \rsqE{-0.235} & \rsqE{< 0}& \rsqD{+0.498} & \rsqD{+0.498} & \rsqC{+0.645} & \rsqC{+0.645} \\
a\_dur           & \rsqC{+0.538} & \rsqC{+0.647} & \rsqC{+0.647} & \rsqE{-0.170} & \rsqE{< 0} & \rsqD{+0.141} & \rsqD{+0.141} & \rsqC{+0.588} & \rsqC{+0.588} \\
a\_post          & \rsqD{+0.235} & \rsqD{+0.471} & \rsqD{+0.471} & \rsqE{< 0}& \rsqE{< 0} & \rsqE{< 0} & \rsqE{< 0} & \rsqD{+0.490} & \rsqD{+0.490} \\
t\_pre           & \rsqD{+0.097} & \rsqD{+0.492} & \rsqD{+0.492} & \rsqE{< 0} & \rsqE{< 0} & \rsqD{+0.127} & \rsqD{+0.127} & \rsqD{+0.412} & \rsqD{+0.412} \\
t\_dur           & \rsqD{+0.071} & \rsqD{+0.457} & \rsqD{+0.457} & \rsqE{<0} & \rsqE{< 0}& \rsqD{+0.286} & \rsqD{+0.286} & \rsqD{+0.337} & \rsqD{+0.337} \\
t\_post          & \rsqD{+0.111} & \rsqC{+0.645} & \rsqC{+0.645} & \rsqD{+0.173} & \rsqE{< 0} & \rsqD{+0.412} & \rsqD{+0.412} & \rsqD{+0.506} & \rsqD{+0.506} \\
lat\_rate\_mean  & \rsqB{+0.811} & \rsqB{+0.902} & \rsqB{+0.902} & \rsqC{+0.638} & \rsqE{< 0} & \rsqE{< 0} & \rsqE{< 0} & \rsqB{+0.870} & \rsqB{+0.870} \\
\bottomrule
\end{tabular}
\end{table*}

Trajectory-level probes further  show that the full output remains recoverable at every decoder layer ($R^2 \in [0.99,1.00]$), even when individual physical parameters are not. This suggests that the decoder first preserves maneuver-level information as a whole, then transforms it through distributed spatio-temporal representations before the final layers. Detailed per-scenario probes and random-control checks are reported in Appendix~\ref{app:further_layers}. Taken together, these results indicate that in this case the decoder is best understood as a structured generator.

\subsection{Layer 4: Controllability via Latent Intervention}
\label{sec:results_l4}

\begin{table}[h]
    \centering
    \scriptsize
    \caption{Latent traversal: maximum $|\rho_{k,j}|$ per dimension and primary effect.}
    \label{tab:perturbation}
    \begin{tabular}{l c l}
        \hline
        \textbf{Dim} & $\max_j |\rho_{k,j}|$ & \textbf{Primary effect} \\
        \hline
        $z_0$ & 0.87 & $y_\text{init}$ \\
        $z_4$ & 0.71 & $\Delta v$, $a_\text{dur}$ \\
        $z_5$ & 0.87 & $v_\text{start}$, $v_\text{end}$ \\
        $z_9$ & 0.80 & $y_\text{final}$ \\
        $z_1, z_2, z_3, z_6, z_7, z_8$ & $\leq 0.39$ & weak or none \\
        \hline
    \end{tabular}
\end{table}

\paragraph{Latent Traversal}
Table~\ref{tab:perturbation} reports the maximum $|\rho_{k,j}|$ from latent traversal. Five dimensions exhibit clear monotonic responses ($|\rho| > 0.7$): $z_0$ for $y_\text{init}$, $z_4$ for $\Delta v$ and $a_\text{dur}$, $z_5$ for velocity, and $z_9$ for $y_\text{final}$. These match the axis-aligned dimensions identified in Layer 1, confirming that the statistical association measured by probing corresponds to a coherent directional dependence.

\paragraph{Functional form and practical sensitivity.}
Table~\ref{tab:sensitivity} shows the normalized response $|\Delta \tilde{p}|$ at nudge $\delta = 0.2$. Effects exceeding the practical threshold ($|\Delta \tilde{p}| > 0.1$) concentrate on the five (dimension, parameter) pairs identified above. Most other dimensions are negligibly sensitive, confirming that controllable structure is concentrated in a low-dimensional subset. But identifying \emph{where} response occurs does not tell us \emph{how}
it scales. So we do polynomial fits to the global nudge response (Table~\ref{tab:poly_fit}) shows degree four for all five effective pairs, with $R^2 \in [0.34, 0.77]$. Linear and low-order polynomial fits are inadequate here indicating that the mapping from latent perturbation to physical parameter change is nonlinear. Linear control intuitions, where doubling a perturbation doubles the response, will not hold for the model. Doing analysis per scenario further sharpens the picture. In (Table~\ref{tab:scenario_poly}) the required degree drops to between one and three and varies systematically. For example: Cut out left shows a linear fit for $y_\text{final}$, while cut through scenarios consistently require cubic fits for the same parameter. So while interactive control is possible in principle, we can’t rely on the neat linear scaling that is often assumed in interpretations of disentangled representations in models.

\begin{table}[h]
\centering
\scriptsize
\setlength{\tabcolsep}{4pt}
\caption{Sensitivity at nudge $\delta = 0.2$ in units of population standard deviation.}
\label{tab:sensitivity}
\begin{tabular}{lcccccccccccc l}
\toprule
 & $y_\text{init}$ & $y_\text{final}$ & $t_\text{pre}$ & $t_\text{dur}$ & $t_\text{post}$ & $\bar{\dot{y}}$ & $v_\text{start}$ & $v_\text{end}$ & $\Delta v$ & $a_\text{pre}$ & $a_\text{dur}$ & $a_\text{post}$ & Role \\
\midrule
$z_4$ & 0.01 & 0.01 & 0.00 & 0.00 & 0.00 & 0.00 & 0.05 & 0.05 & \fbox{\textbf{0.14}} & 0.02 & \fbox{\textbf{0.12}} & 0.05 & Long. \\
$z_5$ & 0.02 & 0.03 & 0.01 & 0.02 & 0.02 & 0.01 & \fbox{\textbf{0.16}} & \fbox{\textbf{0.16}} & 0.01 & 0.01 & 0.01 & 0.00 & Long. \\
$z_9$ & 0.04 & \fbox{\textbf{0.11}} & 0.00 & 0.02 & 0.03 & 0.04 & 0.04 & 0.04 & 0.00 & 0.00 & 0.00 & 0.00 & Lat. \\
$z_1$-$z_3$, $z_6$-$z_8$ & \multicolumn{12}{c}{\textit{all entries} $|\Delta \tilde{p}| < 0.1$} & Mixed/weak \\
\bottomrule
\end{tabular}
\end{table}

\begin{table}[h]
    \centering
    \scriptsize

    \begin{minipage}{0.48\linewidth}
        \centering
        \caption{Global polynomial fits for nudge response.}
        \label{tab:poly_fit}
        \begin{tabular}{l l c c}
        \hline
        \textbf{Dim} & \textbf{Parameter} & \textbf{$R^2$} & \textbf{Degree} \\
        \hline
        $z_4$ & $a_\text{dur}$    & 0.34 & $z^4$ \\
        $z_4$ & $\Delta v$        & 0.51 & $z^4$ \\
        $z_5$ & $v_\text{start}$  & 0.76 & $z^4$ \\
        $z_5$ & $v_\text{end}$    & 0.77 & $z^4$ \\
        $z_9$ & $y_\text{final}$  & 0.65 & $z^4$ \\
        \hline
        \end{tabular}
    \end{minipage}
    \hfill
    \begin{minipage}{0.48\linewidth}
        \centering
        \caption{Scenario stratified polynomial degrees.}
        \label{tab:scenario_poly}
        \begin{tabular}{l c c c c c c}
        \hline
        \textbf{Parameter} & \textbf{CTR} & \textbf{CIR} & \textbf{COR} & \textbf{COL} & \textbf{CIL} & \textbf{CTL} \\
        \hline
        $y_\text{final}$   & $z^3$ & $z^2$ & $z^3$ & $z^1$ & $z^3$ & $z^3$ \\
        $v_\text{start}$   & $z^3$ & $z^2$ & $z^1$ & $z^2$ & $z^2$ & $z^3$ \\
        $v_\text{end}$     & $z^1$ & $z^3$ & $z^2$ & $z^2$ & $z^2$ & $z^3$ \\
        $\Delta v$         & $z^3$ & $z^1$ & $z^3$ & $z^3$ & $z^3$ & $z^1$ \\
        Acceleration       & $z^3$ & $z^3$ & $z^3$ & $z^3$ & $z^3$ & $z^2$ \\
        \hline
        \end{tabular}
    \end{minipage}

\end{table}
\subsection{Layer 5: Physical Plausibility}
\label{sec:results_l5}
Table~\ref{tab:pvr_per_scenario_vae} reports the per-scenario pass rates 
on training and generated trajectories, alongside the per-constraint 
pass rates for friction-circle, lateral acceleration, lateral jerk, 
longitudinal acceleration, and lateral velocity. Five of 
six scenarios match training-data validity within $\pm 5$ percentage 
points. In case of CTL it degrades by $-9.8$ , with the largest 
contribution from a $-21.7$-point drop in jerk validity 
($p_{j_y} : 98.3\% \to 76.6\%$). This likely comes from two things. Cut through trajectories are very less in the training data, so the model handles them less smoothly. Also, Jerk is very sensitive, so small errors in velocity get amplified.
 
\begin{table*}[t]
\centering
\caption{Per-scenario pass rates on training and generated trajectories.}
\label{tab:pvr_per_scenario_vae}

\footnotesize
\setlength{\tabcolsep}{3.4pt}
\renewcommand{\arraystretch}{1.02}

\begin{subtable}[t]{0.49\textwidth}
\centering
\begin{tabular}{l c c c c c c c}
\toprule
\textbf{Scen.} & \textbf{Src.} &
$p_f$ & $p_{a_y}$ & $p_{j_y}$ & $p_{a_x}$ & $p_{v_y}$ & \textbf{PVR} \\
\midrule
CIL & train    & 99.9 & 99.9 & 99.5 & 98.8 & 97.3 & 95.9 \\
    & gen      & 99.6 & 97.3 & 92.0 & 99.9 & 96.3 & 90.8 \\
    & $\Delta$ & $-0.3$ & $-2.6$ & $-7.5$ & $+1.2$ & $-1.0$ & $\mathbf{-5.1}$ \\
\addlinespace[2pt]

CIR & train    & 93.1 & 99.3 & 96.5 & 85.9 & 100.0 & 85.3 \\
    & gen      & 93.8 & 99.2 & 97.6 & 89.8 & 99.9 & 89.6 \\
    & $\Delta$ & $+0.7$ & $-0.1$ & $+1.1$ & $+3.9$ & $-0.1$ & $\mathbf{+4.3}$ \\
\addlinespace[2pt]

COL & train    & 99.9 & 99.9 & 99.8 & 99.7 & 99.9 & 99.5 \\
    & gen      & 99.9 & 99.9 & 99.5 & 99.9 & 100.0 & 99.5 \\
    & $\Delta$ & $\sim 0$ & $-0.1$ & $-0.3$ & $+0.2$ & $\sim 0$ & $\mathbf{\sim 0}$ \\
\bottomrule
\end{tabular}
\end{subtable}
\hfill
\begin{subtable}[t]{0.49\textwidth}
\centering
\begin{tabular}{l c c c c c c c}
\toprule
\textbf{Scen.} & \textbf{Src.} &
$p_f$ & $p_{a_y}$ & $p_{j_y}$ & $p_{a_x}$ & $p_{v_y}$ & \textbf{PVR} \\
\midrule
COR & train    & 96.8 & 99.6 & 98.6 & 92.3 & 95.7 & 88.3 \\
    & gen      & 98.3 & 99.1 & 97.2 & 96.7 & 97.0 & 92.8 \\
    & $\Delta$ & $+1.5$ & $-0.5$ & $-1.4$ & $+4.4$ & $+1.3$ & $\mathbf{+4.5}$ \\
\addlinespace[2pt]

CTL & train    & 99.0 & 99.3 & 98.3 & 95.9 & 87.8 & 84.3 \\
    & gen      & 94.9 & 86.4 & 76.6 & 98.8 & 88.0 & 74.5 \\
    & $\Delta$ & $-4.1$ & $-12.9$ & $-21.7$ & $+3.0$ & $+0.2$ & $\mathbf{-9.8}$ \\
\addlinespace[2pt]

CTR & train    & 89.6 & 98.8 & 95.7 & 81.3 & 93.9 & 76.7 \\
    & gen      & 88.2 & 97.2 & 91.4 & 81.8 & 94.5 & 78.5 \\
    & $\Delta$ & $-1.5$ & $-1.6$ & $-4.3$ & $+0.6$ & $+0.6$ & $\mathbf{+1.8}$ \\
\bottomrule
\end{tabular}
\end{subtable}

\end{table*}

\section{Discussion}

The five-layer protocol reveals behaviors that output-based evaluations alone would miss. Although standard evaluations can suggest competitive generation performance, they do not show whether the model learns transparent, controllable, or physically valid representations. Layer~1 shows partial alignment between learned representations and human-defined kinematic descriptors. This is not necessarily a failure, since kinematic descriptors often assume linear relationships while neural networks encode driving behavior nonlinearly and across multiple dimensions. However, partial alignment makes intervention harder, because target quantities may not correspond to clean latent directions. The statistical baseline comparison in Layer~2 suggests that the network learns additional structure, but remains biased by patterns already present in the dataset. Layer~3 further shows that network layers are not complete black boxes. They can be interpreted and that  information is reorganized across decoder depth, so interventions may be more effective at the layers where target properties are constructed rather than only at the latent input. Layer~4 builds from Layer~1 and 2 showing some physical quantities are linearly controllable, while others require nonlinear or scenario-dependent mappings, making latent steering unreliable. Layer~5 shows that high output quality does not guarantee physical consistency, motivating physics-informed training in the future. Overall, the protocol improves transparency by showing what the model learns, how information is organized, where interventions should be applied, and whether generated scenarios remain physically executable.

\noindent\textit{Note:} We provide a detailed comparison of the proposed protocol with two state-of-the-art benchmarks, WOSAC and Waymo E2E, in Section~\ref{sec:sota_comparison}.

\section{Conclusion}
\label{sec:conclusion}

We introduced a five-layer protocol for evaluating generative scenario models beyond output quality alone. The protocol examines whether models learn meaningful driving structure, respond controllably to interventions, and generate physically executable trajectories. When applied to baseline architectures trained on highway maneuver data, it revealed important behaviors that standard metrics miss, including partial kinematic alignment, nonlinear and scenario-dependent control responses, and physically invalid outputs. By making these hidden model behaviors visible, the protocol improves transparency and helps build trust for industrial adoption. These findings suggest that future generative driving models should be evaluated across structural, controllability, and physical dimensions, and should be designed with stronger kinematic supervision, scenario-calibrated control mechanisms, layer-specific interventions, and physics-informed training objectives.
\bibliographystyle{unsrtnat}
\bibliography{References}

\begin{thebibliography}{41}
\providecommand{\natexlab}[1]{#1}
\providecommand{\url}[1]{\texttt{#1}}
\expandafter\ifx\csname urlstyle\endcsname\relax
  \providecommand{\doi}[1]{doi: #1}\else
  \providecommand{\doi}{doi: \begingroup \urlstyle{rm}\Url}\fi

\bibitem[Hu et~al.(2023)Hu, Yang, Chen, Li, Sima, Zhu, Chai, Du, Lin, Wang,
  et~al.]{hu2023planning}
Yihan Hu, Jiazhi Yang, Li~Chen, Keyu Li, Chonghao Sima, Xizhou Zhu, Siqi Chai,
  Senyao Du, Tianwei Lin, Wenhai Wang, et~al.
\newblock Planning-oriented autonomous driving.
\newblock In \emph{Proceedings of the IEEE/CVF conference on computer vision
  and pattern recognition}, pages 17853--17862, 2023.

\bibitem[Lee and Kwon(2022)]{lee2022adas}
Dongsu Lee and Minhae Kwon.
\newblock Adas-rl: Safety learning approach for stable autonomous driving.
\newblock \emph{ICT Express}, 8\penalty0 (3):\penalty0 479--483, 2022.

\bibitem[Koopman and Wagner(2017)]{koopman2017autonomous}
Philip Koopman and Michael Wagner.
\newblock Autonomous vehicle safety: An interdisciplinary challenge.
\newblock \emph{IEEE Intelligent Transportation Systems Magazine}, 9\penalty0
  (1):\penalty0 90--96, 2017.

\bibitem[Zhong et~al.(2021)Zhong, Tang, Zhou, Neves, Liu, and
  Ray]{zhong2021survey}
Ziyuan Zhong, Yun Tang, Yuan Zhou, Vania de~Oliveira Neves, Yang Liu, and
  Baishakhi Ray.
\newblock A survey on scenario-based testing for automated driving systems in
  high-fidelity simulation.
\newblock \emph{arXiv preprint arXiv:2112.00964}, 2021.

\bibitem[Menzel et~al.(2018)Menzel, Bagschik, and Maurer]{menzel2018scenarios}
Till Menzel, Gerrit Bagschik, and Markus Maurer.
\newblock Scenarios for development, test and validation of automated vehicles.
\newblock In \emph{2018 IEEE intelligent vehicles symposium (IV)}, pages
  1821--1827. IEEE, 2018.

\bibitem[Gambi et~al.(2022)Gambi, Nguyen, Ahmed, and
  Fraser]{gambi2022generating}
Alessio Gambi, Vuong Nguyen, Jasim Ahmed, and Gordon Fraser.
\newblock Generating critical driving scenarios from accident sketches.
\newblock In \emph{2022 IEEE International Conference On Artificial
  Intelligence Testing (AITest)}, pages 95--102. IEEE, 2022.

\bibitem[Ding et~al.(2023)Ding, Xu, Arief, Lin, Li, and Zhao]{ding2023survey}
Wenhao Ding, Chejian Xu, Mansur Arief, Haohong Lin, Bo~Li, and Ding Zhao.
\newblock A survey on safety-critical driving scenario generation—a
  methodological perspective.
\newblock \emph{IEEE Transactions on Intelligent Transportation Systems},
  24\penalty0 (7):\penalty0 6971--6988, 2023.

\bibitem[Sauerbier et~al.(2019)Sauerbier, Bock, Weber, and
  Eckstein]{sauerbier2019definition}
Jan Sauerbier, Julian Bock, Hendrik Weber, and Lutz Eckstein.
\newblock Definition of scenarios for safety validation of automated driving
  functions.
\newblock \emph{ATZ worldwide}, 121\penalty0 (1):\penalty0 42--45, 2019.

\bibitem[Krajewski et~al.(2019)Krajewski, Moers, Meister, and
  Eckstein]{krajewski2019beziervae}
Robert Krajewski, Tobias Moers, Adrian Meister, and Lutz Eckstein.
\newblock B{\'e}ziervae: Improved trajectory modeling using variational
  autoencoders for the safety validation of highly automated vehicles.
\newblock In \emph{2019 IEEE intelligent transportation systems conference
  (ITSC)}, pages 3788--3795. IEEE, 2019.

\bibitem[Chang et~al.(2024)Chang, Wang, Zhang, Ge, and
  Li]{chang2024llmscenario}
Cheng Chang, Siqi Wang, Jiawei Zhang, Jingwei Ge, and Li~Li.
\newblock Llmscenario: Large language model driven scenario generation.
\newblock \emph{IEEE Transactions on Systems, Man, and Cybernetics: Systems},
  54\penalty0 (11):\penalty0 6581--6594, 2024.

\bibitem[Lu et~al.(2024)Lu, Ma, Dai, Wang, and Feng]{lu2024realistic}
Qiujing Lu, Meng Ma, Ximiao Dai, Xuanhan Wang, and Shuo Feng.
\newblock Realistic corner case generation for autonomous vehicles with
  multimodal large language model.
\newblock \emph{arXiv preprint arXiv:2412.00243}, 2024.

\bibitem[Xu et~al.(2025)Xu, Petiushko, Zhao, and Li]{xu2025diffscene}
Chejian Xu, Aleksandr Petiushko, Ding Zhao, and Bo~Li.
\newblock Diffscene: Diffusion-based safety-critical scenario generation for
  autonomous vehicles.
\newblock In \emph{Proceedings of the AAAI conference on artificial
  intelligence}, volume~39, pages 8797--8805, 2025.

\bibitem[Russell et~al.(2025)Russell, Hu, Bertoni, Fedoseev, Shotton, Arani,
  and Corrado]{russell2025gaia}
Lloyd Russell, Anthony Hu, Lorenzo Bertoni, George Fedoseev, Jamie Shotton,
  Elahe Arani, and Gianluca Corrado.
\newblock Gaia-2: A controllable multi-view generative world model for
  autonomous driving.
\newblock \emph{arXiv preprint arXiv:2503.20523}, 2025.

\bibitem[Gao et~al.(2025)Gao, Wang, Li, Long, Yang, and Shen]{gao2025survey}
Haoxiang Gao, Zhongruo Wang, Yaqian Li, Kaiwen Long, Ming Yang, and Yiqing
  Shen.
\newblock A survey for foundation models in autonomous driving.
\newblock In \emph{2025 6th International Conference on Computer Vision and
  Data Mining (ICCVDM)}, pages 63--71. IEEE, 2025.

\bibitem[Alain and Bengio(2016)]{alain2016understanding}
Guillaume Alain and Yoshua Bengio.
\newblock Understanding intermediate layers using linear classifier probes.
\newblock \emph{arXiv preprint arXiv:1610.01644}, 2016.

\bibitem[Fil et~al.(2021)Fil, Mesinovic, Morris, and Wildberger]{fil2021beta}
Miroslav Fil, Munib Mesinovic, Matthew Morris, and Jonas Wildberger.
\newblock beta-vae reproducibility: Challenges and extensions.
\newblock \emph{arXiv preprint arXiv:2112.14278}, 2021.

\bibitem[Olah et~al.(2020)Olah, Cammarata, Schubert, Goh, Petrov, and
  Carter]{olah2020zoom}
Chris Olah, Nick Cammarata, Ludwig Schubert, Gabriel Goh, Michael Petrov, and
  Shan Carter.
\newblock Zoom in: An introduction to circuits.
\newblock \emph{Distill}, 5\penalty0 (3):\penalty0 e00024--001, 2020.

\bibitem[Elhage et~al.(2021)Elhage, Nanda, Olsson, Henighan, Joseph, Mann,
  Askell, Bai, Chen, Conerly, et~al.]{elhage2021mathematical}
Nelson Elhage, Neel Nanda, Catherine Olsson, Tom Henighan, Nicholas Joseph, Ben
  Mann, Amanda Askell, Yuntao Bai, Anna Chen, Tom Conerly, et~al.
\newblock A mathematical framework for transformer circuits.
\newblock \emph{Transformer Circuits Thread}, 1\penalty0 (1):\penalty0 12,
  2021.

\bibitem[Shamir(2004)]{shamir2004should}
Tzila Shamir.
\newblock How should an autonomous vehicle overtake a slower moving vehicle:
  Design and analysis of an optimal trajectory.
\newblock \emph{IEEE Transactions on Automatic Control}, 49\penalty0
  (4):\penalty0 607--610, 2004.

\bibitem[Salvucci(2004)]{salvucci2004inferring}
Dario~D Salvucci.
\newblock Inferring driver intent: A case study in lane-change detection.
\newblock In \emph{Proceedings of the Human Factors and Ergonomics Society
  Annual Meeting}, volume~48, pages 2228--2231. SAGE Publications Sage CA: Los
  Angeles, CA, 2004.

\bibitem[Hoerl and Kennard(1970)]{hoerl1970ridge}
Arthur~E Hoerl and Robert~W Kennard.
\newblock Ridge regression: Biased estimation for nonorthogonal problems.
\newblock \emph{Technometrics}, 12\penalty0 (1):\penalty0 55--67, 1970.

\bibitem[Ma{\'c}kiewicz and Ratajczak(1993)]{mackiewicz1993principal}
Andrzej Ma{\'c}kiewicz and Waldemar Ratajczak.
\newblock Principal components analysis (pca).
\newblock \emph{Computers \& Geosciences}, 19\penalty0 (3):\penalty0 303--342,
  1993.

\bibitem[Burgess et~al.(2018)Burgess, Higgins, Pal, Matthey, Watters,
  Desjardins, and Lerchner]{burgess2018understanding}
Christopher~P. Burgess, Irina Higgins, Arka Pal, Loic Matthey, Nick Watters,
  Guillaume Desjardins, and Alexander Lerchner.
\newblock Understanding disentangling in {$\beta$}-vae.
\newblock \emph{arXiv preprint arXiv:1804.03599}, 2018.

\bibitem[Cohen(1988)]{cohen1988statistical}
Jacob Cohen.
\newblock \emph{Statistical Power Analysis for the Behavioral Sciences}.
\newblock Academic Press, New York, 1988.

\bibitem[Pacejka(2005)]{pacejka2005tire}
Hans Pacejka.
\newblock \emph{Tire and vehicle dynamics}.
\newblock Elsevier, 2005.

\bibitem[de~Gelder et~al.(2022)de~Gelder, Hof, Cator, Paardekooper, den Camp,
  Ploeg, and De~Schutter]{de2022scenario}
Erwin de~Gelder, Jasper Hof, Eric Cator, Jan-Pieter Paardekooper, Olaf~Op den
  Camp, Jeroen Ploeg, and Bart De~Schutter.
\newblock Scenario parameter generation method and scenario representativeness
  metric for scenario-based assessment of automated vehicles.
\newblock \emph{IEEE Transactions on Intelligent Transportation Systems},
  23\penalty0 (10):\penalty0 18794--18807, 2022.

\bibitem[Zofka et~al.(2015)Zofka, Kuhnt, Kohlhaas, Rist, Schamm, and
  Z{\"o}llner]{zofka2015data}
Marc~Ren{\'e} Zofka, Florian Kuhnt, Ralf Kohlhaas, Christoph Rist, Thomas
  Schamm, and J~Marius Z{\"o}llner.
\newblock Data-driven simulation and parametrization of traffic scenarios for
  the development of advanced driver assistance systems.
\newblock In \emph{2015 18th international conference on information fusion
  (FUSION)}, pages 1422--1428. IEEE, 2015.

\bibitem[Thal et~al.(2020)Thal, Znamiec, Henze, Nakamura, Imanaga,
  Antona-Makoshi, Uchida, and Taniguchi]{thal2020incorporating}
Silvia Thal, Holger Znamiec, Roman Henze, Hiroki Nakamura, Hisashi Imanaga,
  Jacobo Antona-Makoshi, Nobuyuki Uchida, and Satoshi Taniguchi.
\newblock Incorporating safety relevance and realistic parameter combinations
  in test-case generation for automated driving safety assessment.
\newblock In \emph{2020 IEEE 23rd International Conference on Intelligent
  Transportation Systems (ITSC)}, pages 1--6. IEEE, 2020.

\bibitem[Kayatas et~al.(2023)Kayatas, Bestle, Bestle, and
  Reick]{kayatas2023generation}
Zafer Kayatas, Dieter Bestle, Pascal Bestle, and Robin Reick.
\newblock Generation of realistic cut-in maneuvers to support safety assessment
  of advanced driver assistance systems.
\newblock \emph{Applied Mechanics}, 4\penalty0 (4):\penalty0 1066--1077, 2023.

\bibitem[Demetriou et~al.(2020)Demetriou, Allsv{\aa}g, Rahrovani, and
  Chehreghani]{demetriou2020generation}
Andreas Demetriou, Henrik Allsv{\aa}g, Sadegh Rahrovani, and Morteza~Haghir
  Chehreghani.
\newblock Generation of driving scenario trajectories with generative
  adversarial networks.
\newblock In \emph{2020 IEEE 23rd international conference on intelligent
  transportation systems (ITSC)}, pages 1--6. IEEE, 2020.

\bibitem[Zhong et~al.(2023)Zhong, Rempe, Xu, Chen, Veer, Che, Ray, and
  Pavone]{zhong2023guided}
Ziyuan Zhong, Davis Rempe, Danfei Xu, Yuxiao Chen, Sushant Veer, Tong Che,
  Baishakhi Ray, and Marco Pavone.
\newblock Guided conditional diffusion for controllable traffic simulation.
\newblock In \emph{2023 IEEE international conference on robotics and
  automation (ICRA)}, pages 3560--3566. IEEE, 2023.

\bibitem[Wang et~al.(2024)Wang, Zhu, Huang, Chen, Zhu, and
  Lu]{wang2024drivedreamer}
Xiaofeng Wang, Zheng Zhu, Guan Huang, Xinze Chen, Jiagang Zhu, and Jiwen Lu.
\newblock Drivedreamer: Towards real-world-drive world models for autonomous
  driving.
\newblock In \emph{European conference on computer vision}, pages 55--72.
  Springer, 2024.

\bibitem[Zhou et~al.(2025)Zhou, Cai, Zhao, Zhang, Huang, Zhou, and
  Ma]{zhou2025autovla}
Zewei Zhou, Tianhui Cai, Seth~Z Zhao, Yun Zhang, Zhiyu Huang, Bolei Zhou, and
  Jiaqi Ma.
\newblock Autovla: A vision-language-action model for end-to-end autonomous
  driving with adaptive reasoning and reinforcement fine-tuning.
\newblock \emph{arXiv preprint arXiv:2506.13757}, 2025.

\bibitem[Wang et~al.(2025{\natexlab{a}})Wang, Sun, Li, Wang, Peng, Ye,
  Zarrouki, Li, Piccinini, Xie, et~al.]{wang2025enhancing}
Dingrui Wang, Zhexiao Sun, Zhouheng Li, Cheng Wang, Youlun Peng, Hongyuan Ye,
  Baha Zarrouki, Wei Li, Mattia Piccinini, Lei Xie, et~al.
\newblock Enhancing physical consistency in lightweight world models.
\newblock \emph{arXiv preprint arXiv:2509.12437}, 2025{\natexlab{a}}.

\bibitem[Wang et~al.(2025{\natexlab{b}})Wang, Ye, Liang, Sun, Lu, Zhang, Zhao,
  Gao, Seegert, Sch{\"a}fer, et~al.]{wang2025target}
Dingrui Wang, Hongyuan Ye, Zhihao Liang, Zhexiao Sun, Zhaowei Lu, Yuchen Zhang,
  Yuyu Zhao, Yuan Gao, Marvin Seegert, Finn Sch{\"a}fer, et~al.
\newblock Target-bench: Can world models achieve mapless path planning with
  semantic targets?
\newblock \emph{arXiv preprint arXiv:2511.17792}, 2025{\natexlab{b}}.

\bibitem[Chen et~al.(2018)Chen, Li, Grosse, and Duvenaud]{chen2018isolating}
Ricky~TQ Chen, Xuechen Li, Roger~B Grosse, and David~K Duvenaud.
\newblock Isolating sources of disentanglement in variational autoencoders.
\newblock \emph{Advances in neural information processing systems}, 31, 2018.

\bibitem[Eastwood and Williams(2018)]{eastwood2018framework}
Cian Eastwood and Christopher~KI Williams.
\newblock A framework for the quantitative evaluation of disentangled
  representations.
\newblock In \emph{International conference on learning representations}, 2018.

\bibitem[Atakishiyev et~al.(2024)Atakishiyev, Salameh, Yao, and
  Goebel]{atakishiyev2024explainable}
Shahin Atakishiyev, Mohammad Salameh, Hengshuai Yao, and Randy Goebel.
\newblock Explainable artificial intelligence for autonomous driving: A
  comprehensive overview and field guide for future research directions.
\newblock \emph{IEEE Access}, 12:\penalty0 101603--101625, 2024.

\bibitem[Itkina and Kochenderfer(2023)]{itkina2023interpretable}
Masha Itkina and Mykel Kochenderfer.
\newblock Interpretable self-aware neural networks for robust trajectory
  prediction.
\newblock In \emph{Conference on Robot Learning}, pages 606--617. PMLR, 2023.

\bibitem[Neumeier et~al.(2021)Neumeier, Botsch, Tollk{\"u}hn, and
  Berberich]{neumeier2021variational}
Marion Neumeier, Michael Botsch, Andreas Tollk{\"u}hn, and Thomas Berberich.
\newblock Variational autoencoder-based vehicle trajectory prediction with an
  interpretable latent space.
\newblock In \emph{2021 IEEE International Intelligent Transportation Systems
  Conference (ITSC)}, pages 820--827. IEEE, 2021.

\bibitem[Kingma et~al.(2019)Kingma, Welling, et~al.]{kingma2019introduction}
Diederik~P Kingma, Max Welling, et~al.
\newblock An introduction to variational autoencoders.
\newblock \emph{Foundations and Trends in Machine Learning}, 12\penalty0
  (4):\penalty0 307--392, 2019.

\end{thebibliography}

\appendix

\section{Technical appendices and supplementary material}
\subsection{Related Work}
\label{app:related_work}

Scenario-based testing has traditionally relied on parameterized models that encode kinematic properties through expert-defined parameter sets~\cite{de2022scenario,zofka2015data,thal2020incorporating}. While interpretable, these approaches are labor-intensive to extend and struggle to cover rare scenarios. Learning-based methods address this limitation by capturing complex interaction patterns directly from data. Examples include Variational Autoencoders~\cite{kayatas2023generation,krajewski2019beziervae}, Generative Adversarial Networks~\cite{demetriou2020generation}, large language models~\cite{chang2024llmscenario,lu2024realistic}, diffusion models~\cite{xu2025diffscene,zhong2023guided}, and world models~\cite{wang2024drivedreamer}. These methods can produce diverse and realistic scenarios, but they typically encode driving behavior in high-dimensional latent spaces whose individual dimensions do not correspond to recognizable kinematic quantities. This lack of semantic alignment complicates targeted generation, physical verification, and integration with validation pipelines.

Because of this emphasis on output realism, current evaluation practice for generative scenario models has largely focused on output-level metrics. Reconstruction error and distributional similarity are widely used to assess fidelity, while displacement-based metrics such as ADE and FDE measure trajectory accuracy~\cite{zhou2025autovla}. Some recent work further assesses physical plausibility by comparing distributions of acceleration and jerk~\cite{wang2025enhancing,wang2025target}. However, such comparisons measure closeness to a reference distribution rather than executability under tire friction, actuator, and stability limits. More broadly, evaluation practices remain fragmented, with limited evidence linking generation quality to ADAS validation outcomes~\cite{gao2025survey}. As a result, the internal representations that give rise to generated scenarios are rarely examined, leaving controllability, traceability, and physical grounding largely unaddressed.

This gap connects generative scenario testing to the broader literature on model interpretability. Interpretability research on generative models has primarily concentrated on image synthesis and language modeling. Latent traversal~\cite{fil2021beta} and linear probing~\cite{alain2016understanding} are standard tools for examining whether latent dimensions encode semantically meaningful attributes, while disentanglement metrics quantify the independence of generative factors~\cite{chen2018isolating,eastwood2018framework}. Mechanistic interpretability methods have also begun to expose how specific behaviors emerge from network components~\cite{olah2020zoom,elhage2021mathematical}, although this work has been developed largely for transformer-based language and vision models. Within autonomous driving, interpretability efforts have focused mainly on perception and decision making rather than generative components~\cite{atakishiyev2024explainable}, with some work examining latent representations of maneuver prediction models in discriminative settings~\cite{itkina2023interpretable,neumeier2021variational}. To the best of our knowledge, a systematic interpretability analysis of generative maneuver models that is grounded in physically meaningful kinematic parameters and combined with output-level physical plausibility checks has not yet been undertaken.
\subsection{Mathematical Model}
\label{app:mathematical_model}
A central question in this work is to determine whether the latent 
space of a data-driven generative model encodes physically meaningful 
structure. To investigate this, we need a reference representation 
that assigns concrete physical meaning to maneuver properties 
something that data-driven representations do not provide by construction. 
The parametric model defined in this section serves precisely this 
role: given any maneuver produced by the generative model, the 
parametric model extracts a compact set of physically interpretable 
quantities that can then be correlated against the latent 
representation.

\begin{figure}[h]
    \centering
    \includegraphics[width=0.65\linewidth]{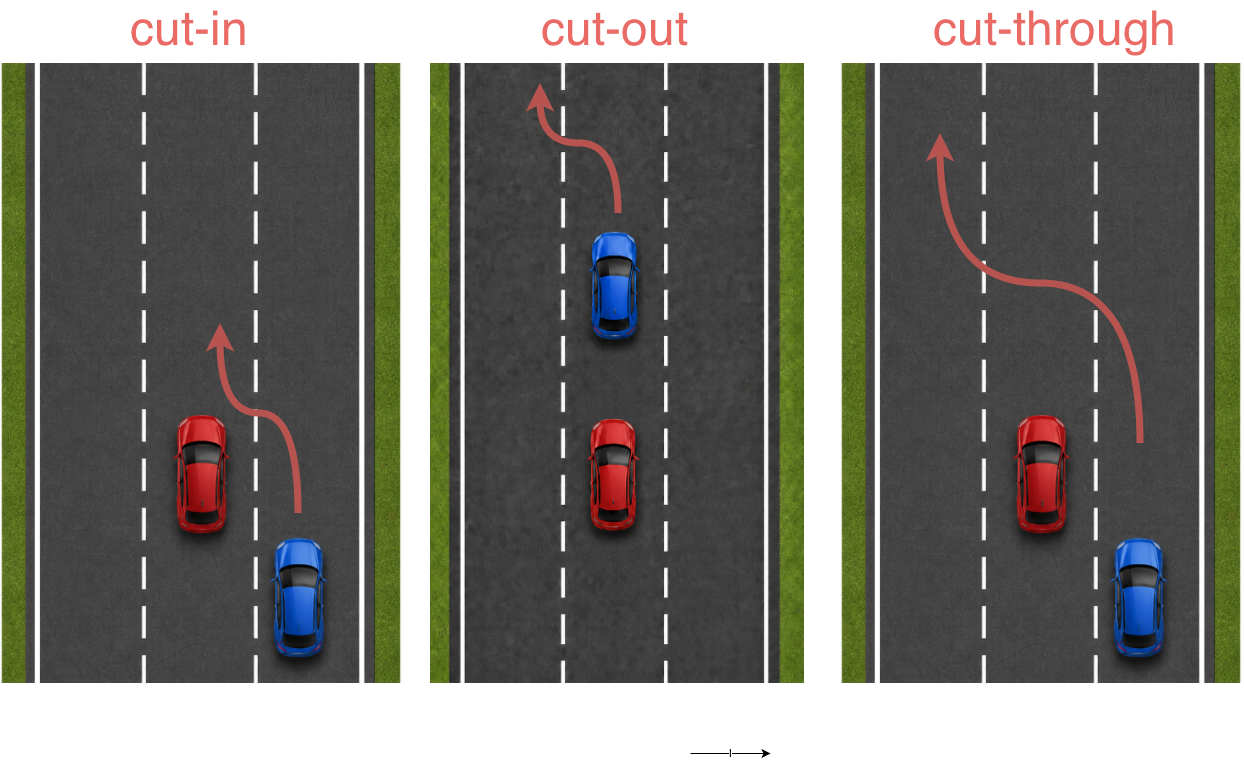}
    \caption{ Overview of the considered scenario types.}
    \label{fig:scenarios_new}
\end{figure}

We examine three highway interaction scenarios commonly 
used in ADAS validation: cut-in, cut-out, and cut-through. In 
each scenario, an actor vehicle executes a lateral maneuver 
relative to an ego vehicle on a highway as shown in Figure~\ref{fig:scenarios_new}. In a cut-in, an actor vehicle merges 
into the ego lane from an adjacent lane, requiring the AV to 
respond to a sudden reduction in headway. In a cut-out, a vehicle 
ahead leaves the ego lane, abruptly exposing slower traffic or a 
stationary obstacle. In a cut-through, the actor crosses the ego 
lane entirely before returning to its original lane, introducing 
more complex lateral dynamics. Together, these they
represent common yet safety-critical highway interactions 
relevant to AV validation.

These highway maneuvers have an inherent temporal structure that 
naturally divides into three causally distinct phases, analogous 
to the preparation, execution, and stabilization structure 
identified in lane-change maneuver models \cite{shamir2004should}. The \textbf{pre-maneuver phase} captures approach and preparation. 
Before initiating a lane change, the actor vehicle establishes its 
conditions: closing or maintaining a gap, adjusting speed, and 
positioning laterally. For example, in a cut-in scenario, this is the phase where 
the actor approaches the ego vehicle from an adjacent lane. Lateral 
position remains stable during this phase, but longitudinal dynamics 
may already be changing with intent. 

The \textbf{active maneuver phase} is the execution window in which 
lateral displacement occurs. Empirical studies of lane-change 
kinematics consistently show that drivers apply progressive steering 
input, producing an S-shaped lateral displacement profile 
\cite{salvucci2004inferring}. A sigmoid function captures this shape 
exactly and is therefore a physically motivated choice rather than 
a convenience approximation. This phase carries the safety-critical 
content of the maneuver: lateral rate, time-to-collision, and 
relative velocity all reach their most critical values here.

The \textbf{post-maneuver phase} captures stabilization. After 
completing the lane change, the actor settles into the new lane and 
either maintains speed, decelerates, or accelerates depending on 
the scenario type. Including this phase is important because the 
post-maneuver longitudinal behavior affects the sustained risk 
exposure of the ego vehicle after the lateral event has concluded. The model is not used for maneuver generation and 
its sole function is parameter extraction, so its known limitations 
such as piecewise constant accelerations and continuity at 
phase boundaries are irrelevant here.
\subsubsection{Model Definition}
Each maneuver consists of two channels observed over a total time horizon $T$: the lateral position $y(t)$ relative to the ego lane center, and the longitudinal velocity $v(t)$.

We define the time boundaries for the three phases as:
\begin{itemize}
    \item $t_1 = t_\text{pre}$ (end of pre-maneuver phase)
    \item $t_2 = t_\text{pre} + t_\text{dur}$ (end of active manuever phase)
    \item $T = t_2 + t_\text{post}$ (total time horizon)
\end{itemize}

\subsubsection{Lateral Motion Model}
\begin{figure}[h]
    \centering
    \includegraphics[width=0.75\linewidth]{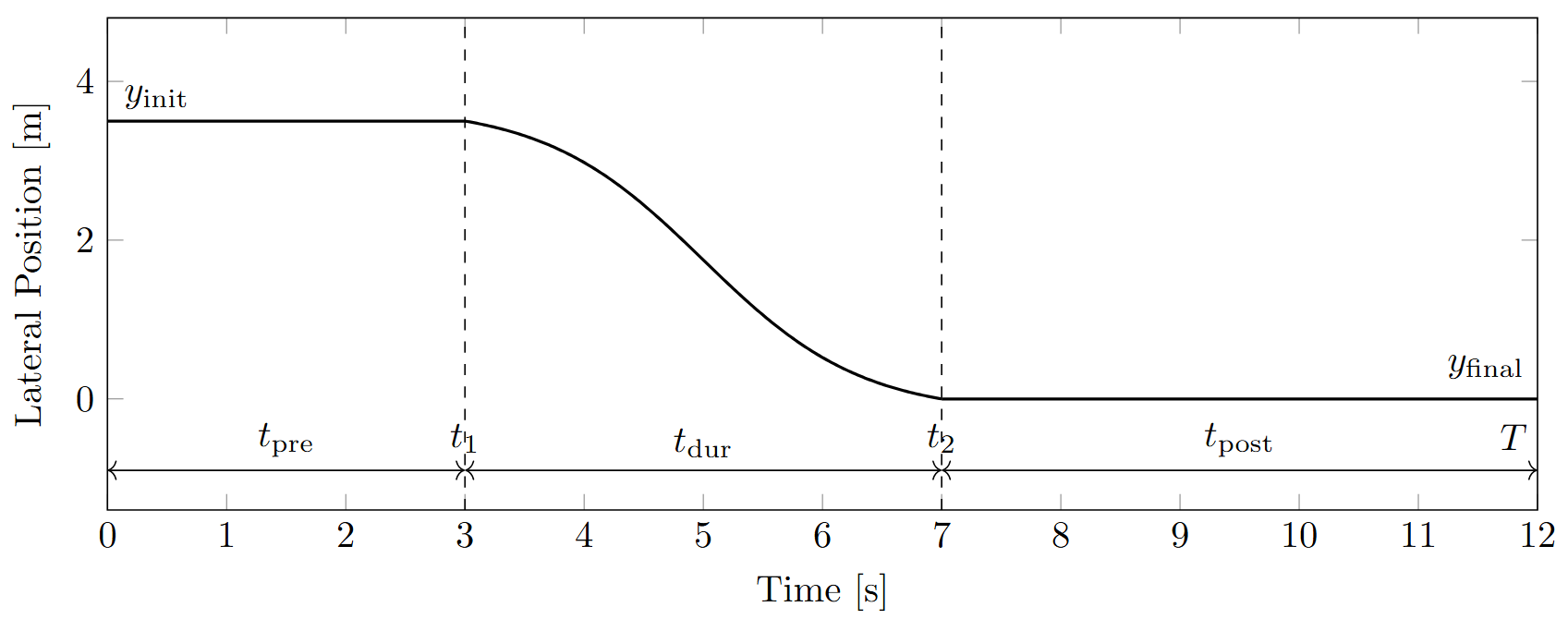}
    \caption{Idealized lateral motion representation.}
    \label{fig:lat}
\end{figure}
The lateral displacement is modeled as a piecewise function. The vehicle maintains its initial lane position during the pre-maneuver phase, executes a smooth sigmoid-based transition during the active maneuver, and holds the final position during the post-maneuver phase. To ensure exact continuity at the phase boundaries, the sigmoid is normalized over the interval $[t_1, t_2]$:

\begin{equation}
\begin{aligned}
y(t) =
\begin{cases}
y_\text{init}, & 0 \le t \le t_1 \\
y_\text{init} + \Delta y \,
\frac{
\sigma(k(t - t_\text{mid})) - \sigma(k(t_1 - t_\text{mid}))
}{
\sigma(k(t_2 - t_\text{mid})) - \sigma(k(t_1 - t_\text{mid}))
},
& t_1 < t \le t_2 \\
y_\text{final}, & t_2 < t \le T
\end{cases}
\end{aligned}
\end{equation}

\noindent where $\sigma(\cdot)$ denotes the logistic sigmoid function

\begin{equation}
    \sigma(x) = \frac{1}{1 + e^{-x}}
    \label{eq:sigmoid}
\end{equation}

\noindent and $t_\text{mid} = t_1 + t_\text{dur}/2$ is the temporal midpoint of the active phase. The steepness parameter $k$ is defined as

\begin{equation}
    k = \frac{6}{t_\text{dur}}
    \label{eq:steepness}
\end{equation}

\noindent This choice places the main portion of the transition within the interval $[t_1,\, t_2]$, since $\sigma(\pm 3) \approx 0.95$ and $\sigma(\mp 3) \approx 0.05$. After normalization, the active-phase segment satisfies the boundary conditions $y(t_1) = y_\text{init}$ and $y(t_2) = y_\text{final}$ exactly.

The mean lateral rate during the active maneuver phase is defined as the total lateral displacement divided by the maneuver duration:

\begin{equation}
    \bar{\dot{y}} = \frac{y_\text{final} - y_\text{init}}{t_\text{dur}}
    \label{eq:lat_rate_mean}
\end{equation}

\noindent This quantity provides a scalar characterization of maneuver aggressiveness: high $|\bar{\dot{y}}|$ indicates a rapid lane change, while low $|\bar{\dot{y}}|$ indicates a gradual one. The sign of $\bar{\dot{y}}$ indicates direction, with positive values corresponding to leftward motion and negative values to rightward motion.

\subsubsection{Longitudinal Velocity Model}
 \begin{figure}[h]
    \centering
    \includegraphics[width=0.75\linewidth]{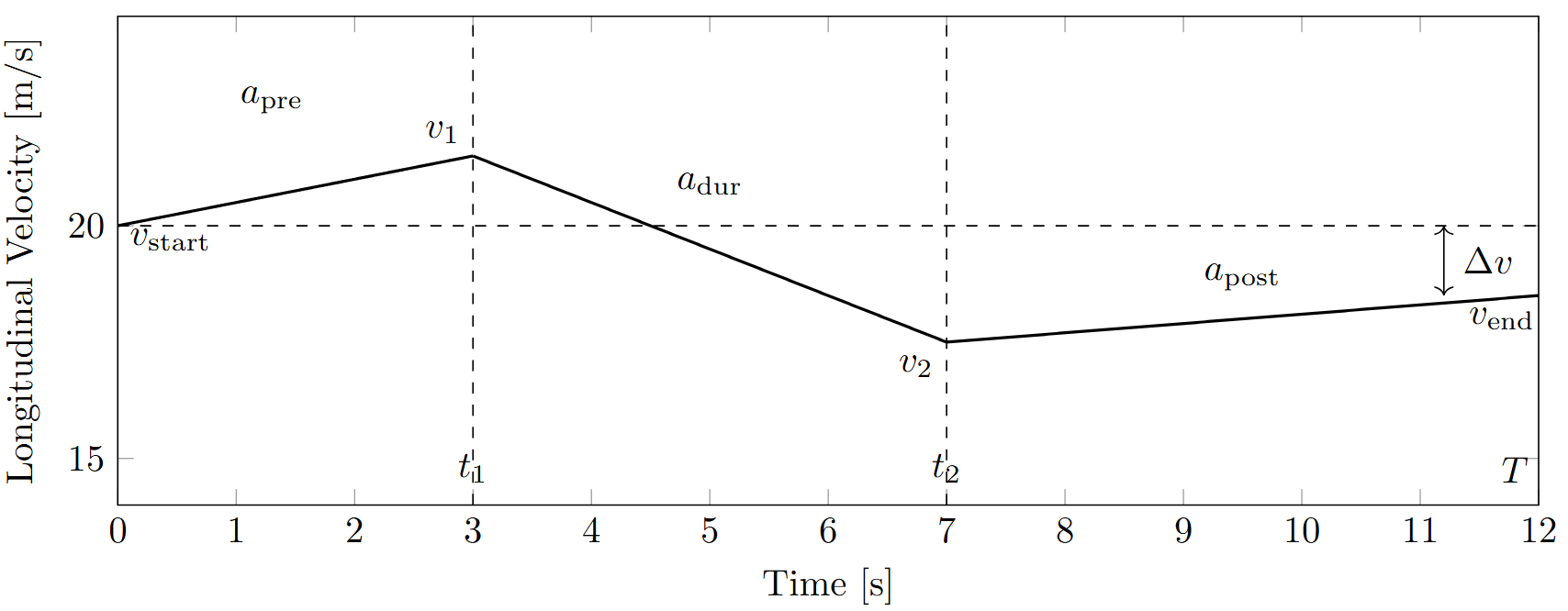}
    \caption{Idealized longitudinal motion representation.}
    \label{fig:vel}
\end{figure}
The longitudinal velocity is modeled using piecewise constant accelerations for each phase:
 
\begin{equation}
    v(t) = 
    \begin{cases} 
        v_\text{start} + a_\text{pre} \cdot t & 0 \le t \le t_1 \\ 
        v_1 + a_\text{dur} \cdot (t - t_1) & t_1 < t \le t_2 \\ 
        v_2 + a_\text{post} \cdot (t - t_2) & t_2 < t \le T 
    \end{cases}
    \label{eq:velocity_piecewise}
\end{equation}
 
\noindent where the boundary velocities are determined by continuity:
 
\begin{align}
    v_1 &= v_\text{start} + a_\text{pre} \cdot t_\text{pre} \label{eq:v1} \\
    v_2 &= v_1 + a_\text{dur} \cdot t_\text{dur} \label{eq:v2}
\end{align}
 
\noindent The final velocity at the end of the maneuver is
 
\begin{equation}
    v_\text{end} = v_2 + a_\text{post} \cdot t_\text{post}
    \label{eq:v_end}
\end{equation}
 
\noindent and the total velocity change over the entire maneuver is
 
\begin{equation}
    \Delta v = v_\text{end} - v_\text{start} = a_\text{pre} \cdot t_\text{pre} + a_\text{dur} \cdot t_\text{dur} + a_\text{post} \cdot t_\text{post}
    \label{eq:delta_v}
\end{equation}
 
\noindent This decomposition captures the common pattern in safety-critical scenarios where velocity changes differ across phases. For example, in a cut-in scenario the actor may approach at constant velocity ($a_\text{pre} \approx 0$), accelerate during the lane change ($a_\text{dur} > 0$), and stabilize afterward ($a_\text{post} \approx 0$).
 
The longitudinal channel exhibits discontinuities in acceleration at the phase boundaries $t_1$ and $t_2$. The velocity profile is continuous ($C^0$) but the acceleration changes instantaneously. For maneuver generation applications requiring $C^2$-continuity, this can be addressed by replacing the piecewise velocity model with a quintic polynomial whose coefficients are determined by boundary conditions on position, velocity, and acceleration, or by introducing jerk-limited trapezoidal ramps at the phase boundaries.
 
In this work, the parametric model serves exclusively for parameter extraction: identifying the physical quantities from measured or generated manuevers. For this purpose, the piecewise formulation is sufficient, as the extracted parameters-positions, velocities, accelerations, and phase durations are determined by the maneuver's macro-structure rather than by the smoothness of transitions at phase boundaries.
 
The maneuver model defines nine independent parameters from which three additional derived quantities are computed. Table~\ref{tab:parameters} summarizes all parameters. The derived quantities are fully determined by the nine independent parameters and do not add degrees of freedom. They are included because they represent physically intuitive maneuver properties, final velocity, velocity change, lane change aggressiveness, that are natural candidates for probing whether the generative model's latent space encodes such concepts.
 
\begin{table}[h]
    \centering
    \caption{Parameters of the multi-phase parametric maneuver model.}
    \label{tab:parameters}
    \begin{tabular}{l l l l}
        \hline
        \textbf{Parameter} & \textbf{Symbol} & \textbf{Unit} & \textbf{Type} \\
        \hline
        Initial lateral position      & $y_\text{init}$       & m       & Independent \\
        Final lateral position        & $y_\text{final}$      & m       & Independent \\
        Initial longitudinal velocity & $v_\text{start}$      & m/s     & Independent \\
        Pre-maneuver acceleration     & $a_\text{pre}$        & m/s$^2$ & Independent \\
        During-maneuver acceleration  & $a_\text{dur}$        & m/s$^2$ & Independent \\
        Post-maneuver acceleration    & $a_\text{post}$       & m/s$^2$ & Independent \\
        Pre-maneuver duration         & $t_\text{pre}$        & s       & Independent \\
        Active maneuver duration      & $t_\text{dur}$        & s       & Independent \\
        Post-maneuver duration        & $t_\text{post}$       & s       & Independent \\
        \hline
        Final velocity                & $v_\text{end}$        & m/s     & Derived (Eq.~\ref{eq:v_end}) \\
        Total velocity change         & $\Delta v$            & m/s     & Derived (Eq.~\ref{eq:delta_v}) \\
        Mean lateral rate             & $\bar{\dot{y}}$       & m/s     & Derived (Eq.~\ref{eq:lat_rate_mean}) \\
        \hline
    \end{tabular}
\end{table}
\subsubsection{Parameter Extraction}
\label{sec:parameter_extraction}

Given a scenario $\{(t_k, y_k, v_k)\}_{k=1}^{N}$, parameter
extraction proceeds in three steps: (i) identification of the phase boundaries
from the lateral channel, (ii) direct readout of geometric parameters, and
(iii) estimation of phase-wise accelerations by least-squares fitting of the
longitudinal channel.

\paragraph{Phase boundary detection.}
The total lateral shift $\Delta y = y_N - y_1$ is evaluated first. If
$|\Delta y| < \epsilon$ with $\epsilon = 0.5\,\text{m}$, the trajectory is
classified as lane-keeping and the time axis is partitioned into equal
thirds. Otherwise, the active-phase boundaries are located as the first time
indices at which the lateral position has covered $5\%$ and $95\%$ of the
total shift:
\begin{equation}
    t_1 = \min\{\, t_k \,:\, \mathrm{sgn}(\Delta y)\,(y_k - y_1) \geq 0.05\,|\Delta y|\,\},
\end{equation}
\begin{equation}
    t_2 = \min\{\, t_k \,:\, \mathrm{sgn}(\Delta y)\,(y_k - y_1) \geq 0.95\,|\Delta y|\,\}.
\end{equation}
The $5\%$-$95\%$ thresholds exclude the near-constant tails of the S-shaped
lateral profile and therefore isolate the segment where lateral motion
actually occurs, without requiring an explicit sigmoid fit. The sign of
$\Delta y$ additionally determines the maneuver direction.

\paragraph{Geometric and timing parameters.}
The initial and final lateral positions are taken directly from the trajectory
endpoints, $y_\text{init} = y_1$ and $y_\text{final} = y_N$. The phase
durations follow from the detected boundaries as
$t_\text{pre} = t_1 - t_0$, $t_\text{dur} = t_2 - t_1$, and
$t_\text{post} = t_N - t_2$.

\paragraph{Longitudinal parameters.}
The initial longitudinal velocity is $v_\text{start} = v_1$. The three
phase-wise accelerations are estimated independently by linear
least-squares regression of velocity against time within each phase:
\begin{equation}
    a_\text{pre}  = \underset{a}{\arg\min} \sum_{t_k \leq t_1} \big(v_k - (v_1 + a\,(t_k - t_0))\big)^2,
\end{equation}
and analogously for $a_\text{dur}$ on $t_1 < t_k \leq t_2$ and $a_\text{post}$
on $t_k > t_2$. This is equivalent to a maximum-likelihood fit of the
piecewise model in Eq.~\eqref{eq:velocity_piecewise} under the assumption of
additive Gaussian noise, and it tolerates the small deviations from perfectly
constant acceleration that occur in measured or generated data.

\paragraph{Derived quantities.}
The derived parameters $v_\text{end}$, $\Delta v$, and $\bar{\dot{y}}$ are
then computed in closed form from the extracted independent parameters using
Eqs.~\eqref{eq:v_end}, \eqref{eq:delta_v}, and \eqref{eq:lat_rate_mean}.

\subsubsection{Analysis of Three-phase Model Fit}

In this work, we use a model from the literature only to demonstrate the probing approach. Since any expert-designed descriptor model introduces its own bias, this bias should be characterized when combining learned representations with expert models. To further elaborate on this, we do the analysis of the three phase model shown in Table~\ref{tab:fit_quality} and find that for lateral position it
moderately fit (mean $R^2 \approx 0.85$), while velocity is systematically
under-fit by piecewise-linear segments (mean $R^2 \approx 0.21$, with some
autocrrelation residuals, $\mathrm{DW} \approx 0$). What it indicates is that piecewise-linear assumption is not the perfect assumption particularly for velocity, where genuine non-linear curvature within phases is being smoothed away rather than fit that is becauseThe model assumes velocity changes linearly within each phase, but real driving produces smooth, curved velocity profiles because acceleration changes gradually. When fitting a straight line, the model captures only the average acceleration, while the true within-phase dynamics (the curvature) are lost and pushed into the residuals. As a result, the parameters are valid summaries, but they do not represent the actual time-varying behavior within each phase. This is one of the main limitation of parametric model.

\begin{table}[h]
\caption{Per-scenario mean fit quality.}
\label{tab:fit_quality}
\centering
\begin{tabular}{llrrrr}
\hline
\textbf{Scenario} & \textbf{Signal} & \textbf{$R^2$} & \textbf{NRMSE} & \textbf{RMSE} & \textbf{DW} \\
\hline
CIL       & $v(t)$ &  0.1113 & 0.2309 & 0.9744 & 0.0103 \\
CIL        & $y(t)$ &  0.8488 & 0.1127 & 0.5467 & 0.0163 \\
CIR      & $v(t)$ &  0.1754 & 0.1982 & 1.1234 & 0.0111 \\
CIR       & $y(t)$ &  0.8257 & 0.1180 & 0.4671 & 0.0177 \\
COL      & $v(t)$ &  0.2157 & 0.2152 & 0.7292 & 0.0112 \\
COL       & $y(t)$ &  0.8332 & 0.1201 & 0.5052 & 0.0181 \\
COR      & $v(t)$ &  0.3288 & 0.1919 & 0.8360 & 0.0131 \\
COR     & $y(t)$ &  0.8216 & 0.1178 & 0.5858 & 0.0173 \\
CTL  & $v(t)$ & -0.1795 & 0.2482 & 1.3190 & 0.0092 \\
CTL  & $y(t)$ &  0.9105 & 0.0821 & 0.8439 & 0.0127 \\
CTR  & $v(t)$ &  0.2820 & 0.1787 & 1.4189 & 0.0116 \\
CTR  & $y(t)$ &  0.9013 & 0.0803 & 0.6715 & 0.0175 \\
\hline
\end{tabular}
\end{table}
\subsection{Dataset- Additional Details}
\label{app:dataset_prep}
The dataset used for training the models is derived from real-world driving data. It combines signals from the ego vehicle’s Electronic Control Unit (ECU) with sensor measurements capturing the surrounding traffic environment. All data are recorded at a sampling rate of 50 Hz. The variables considered include the relative longitudinal and lateral positions \( s(t) \) and \( d(t) \), as well as the absolute speed \( v(t) \) of nearby vehicles.

The dataset includes six types of driving maneuvers that represent common yet safety-critical traffic situations for autonomous vehicle validation. These include cut-in maneuvers to the left and right (CIL, CIR), cut-out maneuvers to the left and right (COL, COR), and cut-through maneuvers to the left and right (CTL, CTR). Cut-in maneuvers occur when a nearby vehicle merges into the ego lane, requiring the autonomous vehicle to respond to sudden intrusions. Cut-out maneuvers involve a vehicle leaving the ego lane unexpectedly, which may create hazardous gaps in traffic, particularly near congestion. Cut-through maneuvers describe situations where a vehicle crosses multiple lanes in a single movement, resulting in more complex lateral interactions (Reference in Figure~\ref{fig:scenarios_new}) 

To construct a labeled dataset for these maneuvers, a two-stage classification approach is applied. The first stage uses rule-based filtering based on the longitudinal distance \( s(t) \) to identify candidate scenarios. The second stage applies a Time-Series Forest (TSF) model, which uses the lateral distance \( d(t) \) to refine and validate the classification results. Prior to classification, preprocessing steps are performed, including removal of missing data, correction for road curvature, and signal smoothing.

The final dataset is organized as refined time-series data consisting of \( N_s \) trajectory samples across all six maneuver types, each observed over \( N_t \) discrete time steps. Every trajectory is associated with a label
\begin{equation}
L^{(i)} \in \{1, 2, 3, 4, 5, 6\},
\end{equation}
where the labels correspond to cut-in left, cut-in right, cut-out left, cut-out right, cut-through left, and cut-through right, respectively.

Each trajectory includes time indices and two motion features: lateral position \( d(t) \) and longitudinal velocity \( v(t) \), sampled at discrete time steps \( t_k \). A single trajectory sample can therefore be written as
\begin{equation}
\left(L^{(i)},\, t_k^{(i)},\, d_k^{(i)},\, v_k^{(i)}\right), \quad i = 1, \dots, N_s,\; k = 1, \dots, N_t.
\end{equation}

For each trajectory segment \( i \), the motion features are combined into a matrix representation, where each column corresponds to a feature (time, lateral position, and velocity):
\begin{equation}
\mathbf{x}^{(i)} =
\begin{bmatrix}
[t_k^{(i)}],\; [d_k^{(i)}],\; [v_k^{(i)}]
\end{bmatrix}, \quad i = 1, \dots, N_s,\; k = 1, \dots, N_t.
\end{equation}

\subsection{Guided VAE}
\label{app:model_details}
This section briefly describes the guided VAE, the generative 
model which is analyzed in this work. A variational autoencoder (VAE) models high-dimensional data by learning a 
probability distribution over a lower-dimensional latent space, enabling the 
generation of diverse and realistic samples~\cite{kingma2019introduction}. 
In this work, the measured manuevers 
\( \mathbf{x} \in \mathbb{R}^{100 \times 3} \) are first encoded into latent 
variables \( \mathbf{z} \in \mathbb{R}^{d} \), which are then decoded to 
reconstruct \( \tilde{\mathbf{x}} \in \mathbb{R}^{100 \times 3} \). 
The encoder network estimates the mean \( \boldsymbol{\mu} \) and variance 
\( \boldsymbol{\sigma}^2 \) of the latent variable vector \( \mathbf{z} \), 
forming an approximate posterior distribution 
\( q_{\boldsymbol{\theta}_E}(\mathbf{z} | \mathbf{x}) \). 
The decoder reconstructs the input by modeling the likelihood 
\( p_{\boldsymbol{\theta}_D}(\tilde{\mathbf{x}} | \mathbf{z}) \) of producing 
an output trajectory \( \tilde{\mathbf{x}} \).

In a standard VAE, the parameters \( \boldsymbol{\theta}_E \) and 
\( \boldsymbol{\theta}_D \) are optimized by minimizing a loss that combines 
a reconstruction term and a regularization term. The reconstruction loss is 
defined as the mean squared error (MSE) between the measured input and its 
reconstruction:
\begin{equation}
\mathrm{MSE}(\tilde{\mathbf{x}}, \mathbf{x}) = \frac{1}{N_s} \sum_{i=1}^{N_s} \| \tilde{\mathbf{x}}^{(i)} - \mathbf{x}^{(i)} \|^2,
\label{eq:MSE}
\end{equation}
while the regularization term minimizes the Kullback-Leibler (KL) divergence 
between the approximate posterior \( q(\mathbf{z}) \) and the multivariate 
Gaussian prior \( p(\mathbf{z}) = 
\mathcal{N}(\boldsymbol{\mu}, \mathrm{diag}(\boldsymbol{\sigma}^2)) \):
\begin{equation}
D_{\mathrm{KL}}\!\left(q(\mathbf{z}) \parallel p(\mathbf{z})\right) = 
\mathbb{E}\!\left[\log \frac{q(\mathbf{z})}{p(\mathbf{z})} \right].
\end{equation}

\begin{figure}[htbp]
    \centering
    \includegraphics[width=0.65\linewidth]{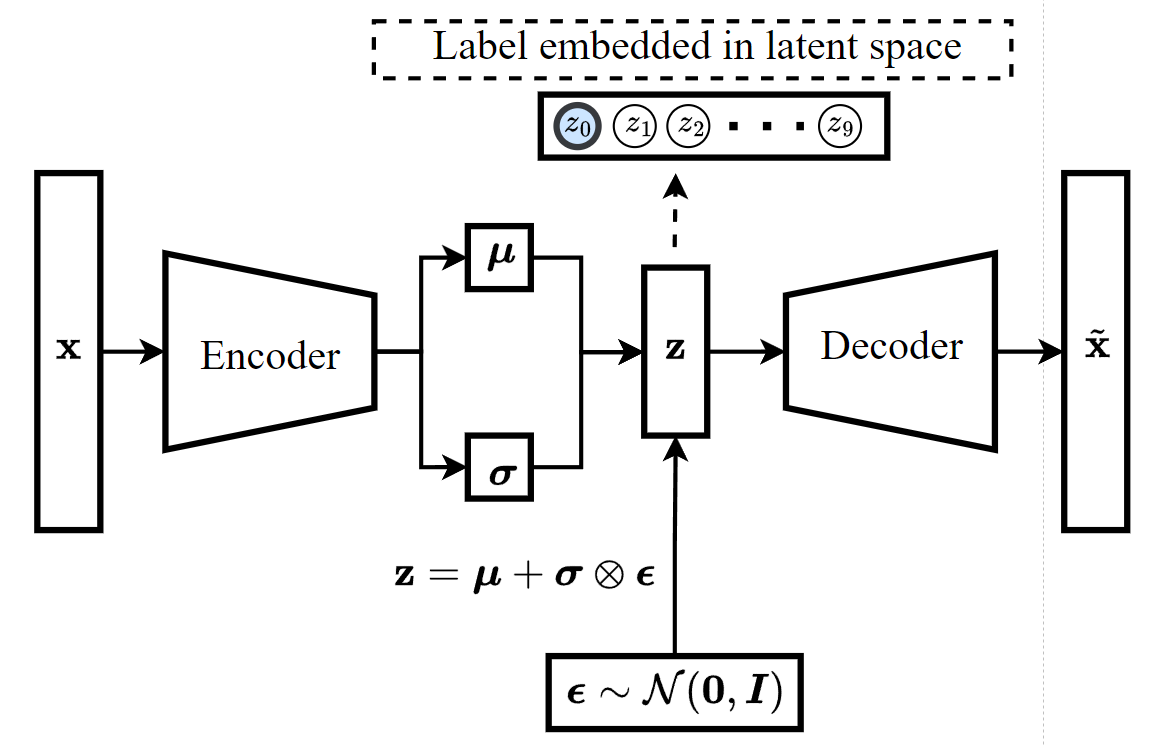}
    \caption{Guided VAE Architecture.}
    \label{fig:architecture}
\end{figure}

However, external conditioning does not shape the latent space 
geometry, which limits both generative coherence and 
controllability. Embedding scenario type directly as a latent 
dimension instead encodes class identity as a continuous geometric 
property of the space, ensuring that sampling naturally respects 
the training distribution, that scenario frequencies are preserved, 
and that the region between class centers remains traversable for 
generating transitional maneuvers. Therefore, we modify the VAE to include scenario 
controllability directly in the latent space, as shown in 
Figure~\ref{fig:architecture}. One latent dimension is assigned 
for scenario representation:
\begin{equation}
    \mathbf{z} = [z_{\text{type}},\, \mathbf{z}_{\text{rest}}],
\end{equation}
where \( z_{\text{type}} \) encodes the scenario type and 
\( \mathbf{z}_{\text{rest}} \) captures the remaining trajectory variability. 
The prior is replaced with a class-conditional Gaussian:
\begin{equation}
    p(\mathbf{z}\mid y) = 
    \mathcal{N}(z_{\text{type}} \mid \mu_y, \sigma_y^2)
    \cdot 
    \mathcal{N}(\mathbf{z}_{\text{rest}} \mid \mathbf{0}, \mathbf{I}),
\end{equation}
where each class \( y \) is assigned a distinct mean \( \mu_y \) along the 
scenario dimension.

To align \( z_{\text{type}} \) with its class center, we introduce an 
auxiliary alignment loss:
\begin{equation}
    \mathcal{L}_{\text{align}} = 
    \lambda_{\text{align}} 
    \,\mathbb{E}_{\mathbf{x}, y}\!\big[(z_{\text{type}} - \mu_y)^2\big].
\end{equation}

The resulting training objective becomes:
\begin{equation}
    \mathcal{L}_{\text{total}} = 
    \mathrm{MSE}(\tilde{\mathbf{x}}, \mathbf{x}) + 
    \beta \, D_{\mathrm{KL}}\!\left(q(\mathbf{z}) \parallel p(\mathbf{z})\right) + 
    \mathcal{L}_{\text{align}},
\label{eq:guided_vae_loss}
\end{equation}
which preserves data fidelity, regularizes the latent distribution, and 
provides explicit control over scenario generation through \( z_{\text{type}} \).

The architecture of the Variational Autoencoder (VAE) used is detailed in Table~\ref{tab:model_archi}. The encoder uses convolutional layers (Conv)  that progressively extract and refine temporal and spatial features from trajectory data. They are followed by a flattening layer (Flatten) and fully connected layers (Dense) that reorder and reduce the representation to a compact latent vector. A Lambda layer is then used to perform sampling from the latent distribution using the reparameterization trick for enabling gradient-based optimization during training. The latent space dimensionality is fixed at \(d=10\). The decoder mirrors the encoder in reverse, a Dense layer expands the latent representation, which is then reshaped and passed through transposed convolutional layers (DeConv) to reconstruct the input sequence. Flatten and Reshape layers are used to reorganize the data.
\begin{table*}[t]
\centering
\caption{Network architecture of the proposed Guided VAE.}
\small
\renewcommand{\arraystretch}{1.3}
\begin{tabular}{|l|c|l|c|}
\hline
\textbf{Encoder Layer} & \textbf{Output Shape} & 
\textbf{Decoder Layer} & \textbf{Output Shape} \\
\hline
Input           & \(100 \times 3\)     & Input           & 10 \\
Conv, ReLU      & \(100 \times 25\)    & Dense, ReLU     & 10000 \\
Conv, ReLU      & \(100 \times 50\)    & Reshape         & \(100 \times 100\) \\
Conv, ReLU      & \(100 \times 100\)   & DeConv, ReLU    & \(100 \times 50\) \\
Flatten         & 10{,}000             & DeConv, ReLU    & \(100 \times 25\) \\
Dense (2x), ReLU & 10                  & Flatten         & 2500 \\
Lambda          & 10                   & Dense           & 300 \\
—               & —                    & Reshape         & \(100 \times 3\) \\
\hline
\end{tabular}

\label{tab:model_archi}
\end{table*}

The implementation uses the following configuration give in Table~\ref{tab:implementation_config}
\begin{table}[t]
\centering
\caption{Implementation configuration and hyperparameters.}
\label{tab:implementation_config}

\begin{tabular}{|p{3.8cm}|p{9.5cm}|}
\hline
\textbf{Setting} & \textbf{Value} \\ \hline

$\beta$ (KL weight) & \texttt{1e-4} \\ \hline

Optimizer & Adam \\ \hline

Initial learning rate & \texttt{1e-3} \\ \hline

Learning-rate schedule & Reduce by \texttt{0.5} after 20 epochs without improvement \\ \hline

Minimum learning rate & \texttt{1e-5} \\ \hline

Maximum epochs & \texttt{500} \\ \hline

Early stopping patience & \texttt{50} epochs \\ \hline

Random seed & \texttt{42} \\ \hline

Five-fold CV & \texttt{KFold(n\_splits=5, shuffle=True, random\_state=0)} \\ \hline

Probe hyperparameters & Ridge: $\alpha = 1.0$; MLP: \texttt{64} hidden units (ReLU, Adam), \texttt{500} max iterations; StandardScaler preprocessing \\ \hline

$\lambda_{\text{align}}$ & Not used (no disentanglement regulariser) \\ \hline

\end{tabular}

\end{table}

\subsubsection{Model Performance}
\label{app:model_performance}

\begin{figure}[h]
    \centering
    \includegraphics[width=0.9\linewidth]{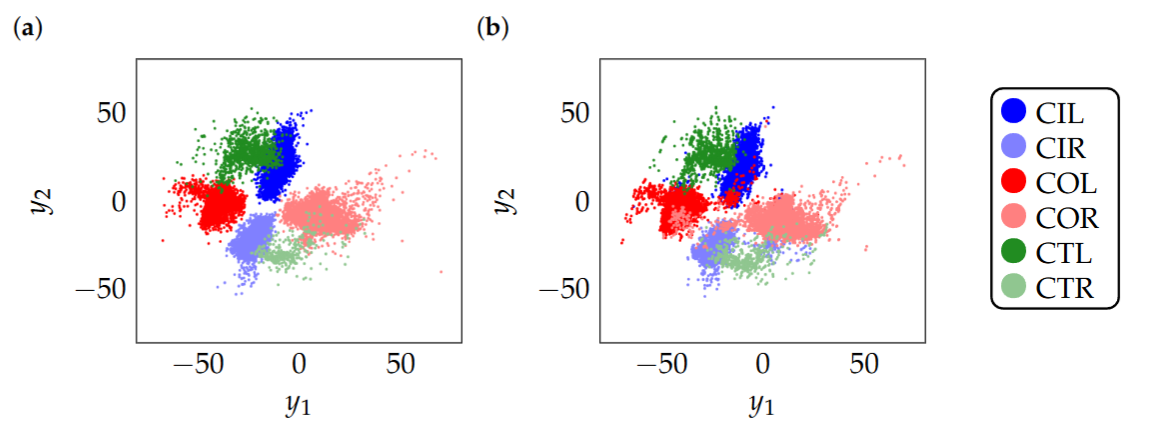}
    \caption{First two modes of PCA applied to (a) measured and (b) generated samples.}
    \label{fig:pca_comparison}
\end{figure}

Reconstruction errors on the validation set remain low across all 
scenario classes, as reported in Table~\ref{tab:combined_mse_error_mean_CE}.. 
The higher MSE observed for CTL is attributable to its smaller 
sample count rather than a structural limitation of the model, 
as scenarios with larger training sets achieve consistently lower 
values.
\begin{table}[h]
\centering
\renewcommand{\arraystretch}{1.25}
\caption{Validation MSE loss and classification error rates (mean values).}
\label{tab:combined_mse_error_mean_CE}

\begin{tabular}{lcc}
\hline
\textbf{Scenario} & \textbf{MSE} & \textbf{Classification Error} \\
\hline
CIL & $5.03 \times 10^{-2}$ & $1.00 \times 10^{-2}$ \\
CIR & $4.17 \times 10^{-2}$ & $2.00 \times 10^{-2}$ \\
COL & $2.05 \times 10^{-2}$ & $1.00 \times 10^{-2}$ \\
COR & $6.41 \times 10^{-2}$ & $1.00 \times 10^{-2}$ \\
CTL & $1.20 \times 10^{-1}$ & $0.00 \times 10^{0}$ \\
CTR & $9.49 \times 10^{-2}$ & $1.00 \times 10^{-2}$ \\
\hline
\end{tabular}
\end{table}
Trajectory-level fidelity is assessed through PCA projections of 
generated and real samples. As shown in 
Figure~\ref{fig:pca_comparison}, the two distributions closely 
overlap, indicating that the global kinematic structure of the 
training data is well preserved. Scenario consistency is further 
assessed by an external classifier, which yields consistently low 
classification errors across all scenario types as reported in 
Table~\ref{tab:combined_mse_error_mean_CE}.

\begin{figure}[h]
    \centering
    \includegraphics[width=0.4\linewidth]{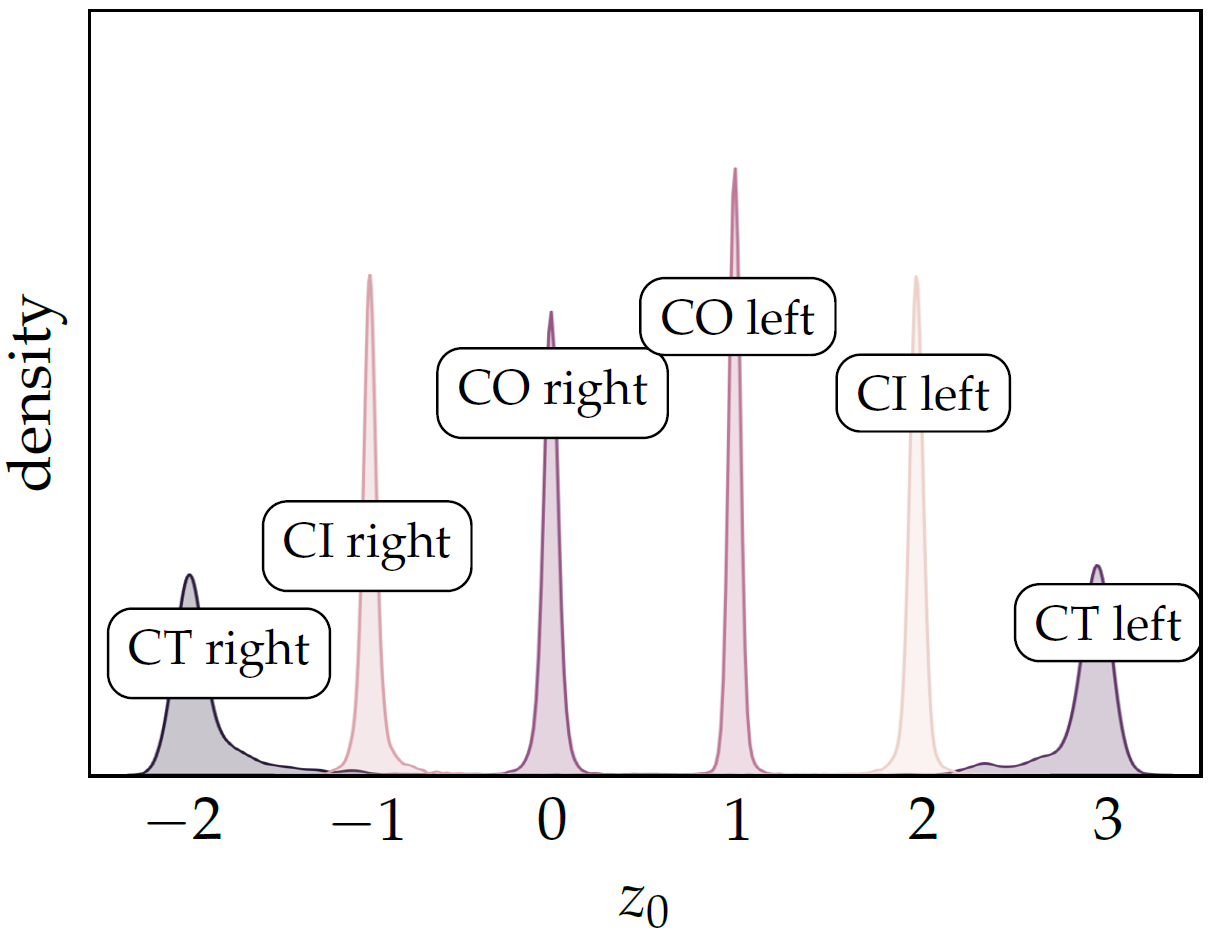}
    \caption{Learned modalities in the latent parameter $z_0$, which separates different scenario types.}
    \label{fig:z0_modalities}
\end{figure}

At the latent level, the alignment loss is verified by inspecting 
the distribution of $z_0$ using KDE. As shown in 
Figure~\ref{fig:z0_modalities}, $z_0$ exhibits a clear multimodal 
distribution with each mode centered around a class-specific 
region, confirming that scenario types are well separated in the 
latent space. The posterior of $z_0$ is approximated as a mixture 
of Gaussians:
\begin{equation}
q(z_0) \approx \sum_{y=0}^{5} \pi_y \cdot 
\mathcal{N}(z_0 \mid \mu_y, \sigma_y^2)
\end{equation}
where $\mu_y$ and $\sigma_y^2$ are the learned mean and variance 
for class $y$, and $\pi_y$ its empirical frequency. Together, 
these results confirm that the model produces realistic 
manuevers with reliable scenario consistency, establishing a 
sound basis for the interpretability analysis that follows.

\subsubsection{Initial Latent Space Analysis}
\begin{figure*}[htbp]
    \centering
    \includegraphics[width=1\textwidth]{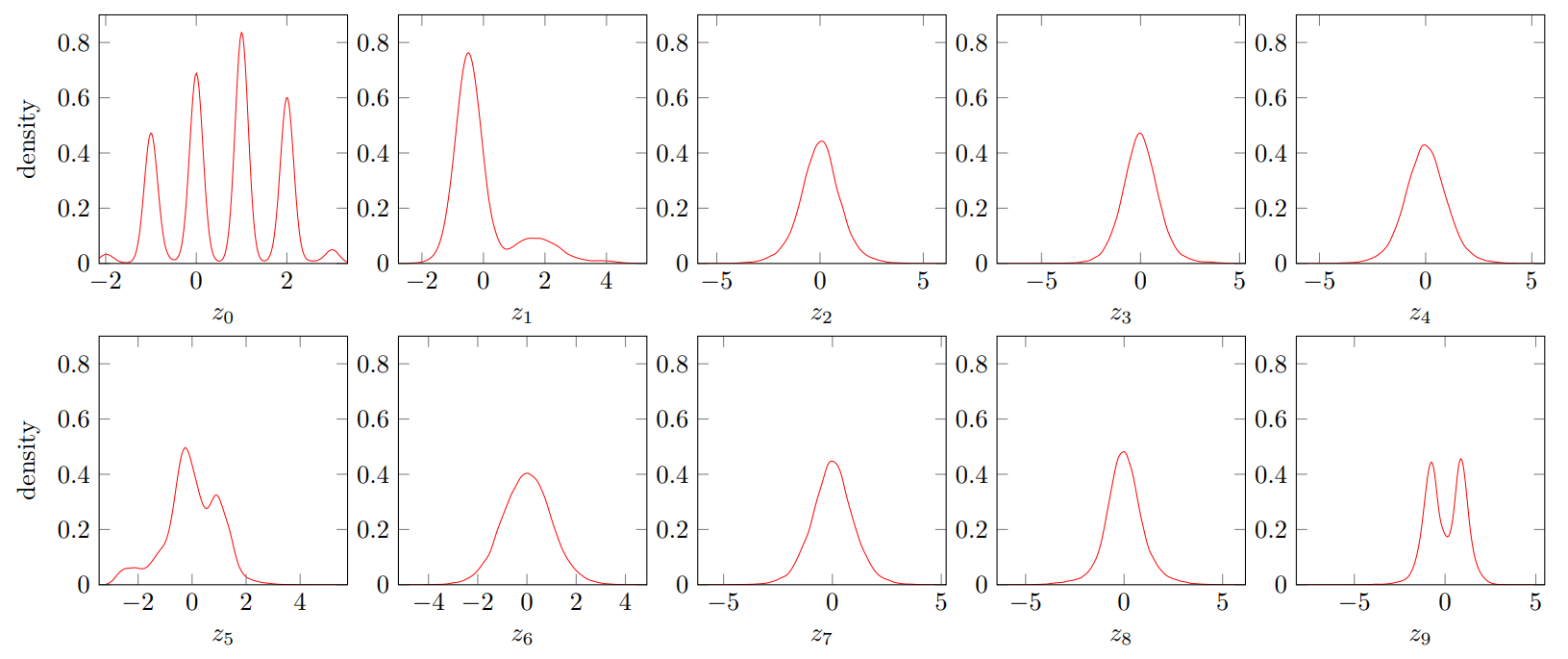}
    \caption{KDE of Latent Parameters.}
    \label{fig:latent}
\end{figure*}

\begin{figure}[h]
    \centering
    \includegraphics[width=0.4\linewidth]{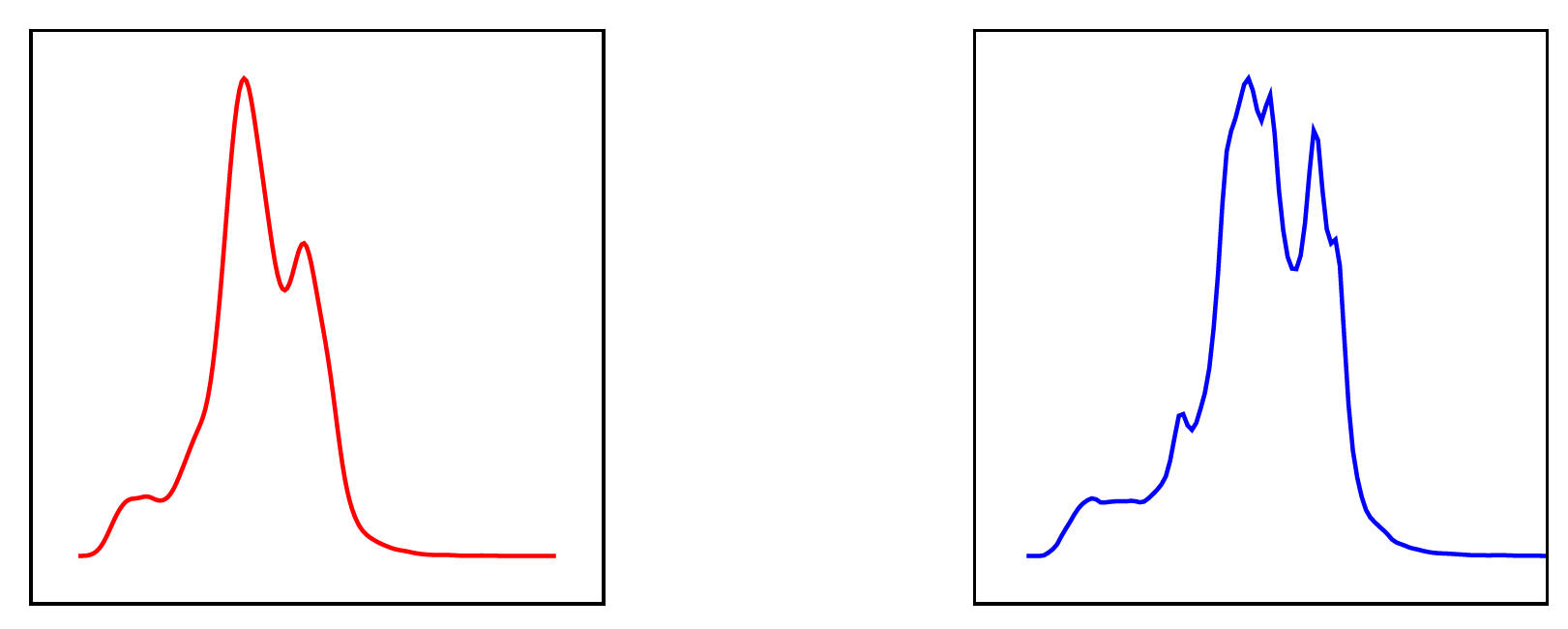}
    \caption{Latent parameter $z_5$ captures generalized velocity behavior. The generated manuevers (red) reproduce the real velocity profiles (blue), including peak dynamics.}
    \label{fig:z6_velocity}
\end{figure}

Building on the latent structure confirmed for $z_0$ , the KDE distributions of 
all latent dimensions are examined as an initial check on whether 
the VAE training objective produced a well-organized latent space. 
Under a standard Gaussian prior, all free dimensions are expected 
to follow unimodal distributions. Deviations from this indicate 
that the encoder has structured those dimensions around specific 
factors of variation in the data, which is a prerequisite for 
physically meaningful latent organization. As shown in 
Figure~\ref{fig:latent}, most free dimensions follow approximately 
Gaussian distributions as expected, however $z_2$, $z_5$, and 
$z_9$ display clear multimodal behavior, suggesting they encode 
structured variation beyond the prior. Notably, $z_5$ shows a 
distribution that closely matches the longitudinal velocity 
profiles of the training data, as shown in 
Figure~\ref{fig:z6_velocity}, indicating that the model 
spontaneously organized this dimension around velocity without 
any explicit supervision. This provides the first concrete 
evidence that the free latent dimensions capture physically 
meaningful properties, which the following analysis investigates 
systematically.
\subsection{Robustness Check in Train/Test Split}

To verify that the performance of guided VAE has no recording level correlation, we additionally evaluated the model under a recording-level split, in which whole recordings were partitioned 70/30 with stratification by dominant scenario so that no recording contributed segments to both sets. As shown in Table~\ref{tab:split_comparison}, mean classification error from $z_0$ remains low under in both cases (1.0\% maneuver-level vs.\ 0.67\% recording-level), confirming that the scenario-encoding latent dimension is robust to the split strategy.

\begin{table}[h]
\centering
\caption{Robustness Check for Train/Test Split}
\renewcommand{\arraystretch}{1.3}
\begin{tabular}{|l|cc|cc|}
\hline
\textbf{Scenario} & \multicolumn{2}{c|}{\textbf{Maneuver-level split}} & \multicolumn{2}{c|}{\textbf{Recording-level split}} \\
\cline{2-5}
 & MSE & Cls. Error & MSE & Cls. Error \\
\hline
CIL  & 5.03e-2 & 1.00e-2 & 1.109e-1 & 1.00e-2 \\
CIR  & 4.17e-2 & 2.00e-2 & 2.68e-2  & 0.00e+0 \\
COL  & 2.05e-2 & 1.00e-2 & 1.796e-1 & 2.00e-2 \\
COR  & 6.41e-2 & 1.00e-2 & 2.627e-1 & 1.00e-2 \\
CTL  & 1.20e-1 & 0.00e+0 & 9.19e-2  & 0.00e+0 \\
CTR  & 9.49e-2 & 1.00e-2 & 1.109e-1 & 0.00e+0 \\
\hline
\end{tabular}

\label{tab:split_comparison}
\end{table}
\subsection{Further Results by Layers}
\label{app:further_layers}
\subsubsection{Layer 1}
\label{app:further_layer_1}
Table~\ref{tab:best_dim} reports, for each physical parameter, the latent 
dimension that best predicts it univariately and the corresponding 
single-dimension $R^2$. The full per-dimension probing matrices are shown 
in Table~\ref{tab:per_dimension_probing}.

\begin{table}[h]
    \centering
    \caption{Per-dimension probing: best latent dimension for each 
    physical parameter and the corresponding univariate $R^2$.}
    \label{tab:best_dim}
  \begin{tabular}{l c c c}
\hline
\textbf{Parameter} & \textbf{Best dim.} & $R^2_\text{best}$ & \textbf{Status} \\
\hline
$y_\text{init}$   & $z_0$ & 0.76 & axis-aligned \\
$y_\text{final}$  & $z_9$ & 0.64 & axis-aligned \\
$v_\text{start}$  & $z_5$ & 0.76 & axis-aligned \\
$v_\text{end}$    & $z_5$ & 0.75 & axis-aligned \\
$\Delta v$        & $z_4$ & 0.51 & axis-aligned \\
$\bar{\dot{y}}$   & $z_0$ & 0.58 & axis-aligned \\
$a_\text{dur}$    & $z_4$ & 0.34 & weak alignment \\
$a_\text{pre}$    & $z_7$ & 0.12 & weak alignment \\
$a_\text{post}$   & $z_7$ & 0.11 & weak alignment \\
$t_\text{pre}$    & $z_1$ & 0.06 & no alignment \\
$t_\text{post}$   & $z_1$ & 0.06 & no alignment \\
$t_\text{dur}$    & $z_5$ & 0.02 & no alignment \\
\hline
\end{tabular}
\end{table}
\begin{table*}[ht]
\centering
\caption{Per-Dimension Probing.}
\label{tab:per_dimension_probing}
\scriptsize
\setlength{\tabcolsep}{4pt}
\renewcommand{\arraystretch}{1.15}
\begin{tabular}{llcccccccccc}
\hline
Type & Variable & z\_0 & z\_1 & z\_2 & z\_3 & z\_4 & z\_5 & z\_6 & z\_7 & z\_8 & z\_9 \\
\hline
Lateral & y\_final & 0.14 & 0.02 & 0.03 & -0.00 & 0.00 & 0.04 & 0.00 & -0.00 & 0.10 & \cellcolor{green!35}0.64 \\
Lateral & y\_init & \cellcolor{green!35}0.76 & 0.00 & 0.01 & 0.03 & 0.01 & 0.02 & 0.07 & 0.01 & 0.01 & 0.12 \\
Lateral & scenario\_category & \cellcolor{green!35}0.75 & 0.00 & -0.00 & -0.00 & 0.00 & 0.00 & 0.01 & 0.00 & 0.03 & 0.00 \\
Lateral & lat\_rate\_mean & \cellcolor{green!35}0.58 & 0.01 & -0.00 & 0.00 & 0.00 & 0.01 & 0.03 & 0.00 & 0.07 & 0.15 \\
Lateral & t\_post & 0.00 & 0.06 & 0.00 & -0.00 & -0.00 & 0.03 & -0.00 & 0.00 & 0.00 & 0.02 \\
Lateral & t\_pre & 0.01 & 0.06 & 0.01 & 0.01 & 0.00 & 0.01 & 0.00 & -0.00 & 0.00 & 0.00 \\
Lateral & t\_dur & 0.02 & 0.02 & 0.01 & 0.01 & 0.00 & 0.02 & -0.00 & -0.00 & 0.00 & 0.00 \\
\hline
Longitudinal & v\_end & 0.00 & 0.00 & -0.00 & -0.00 & 0.06 & \cellcolor{green!35}0.75 & 0.02 & 0.01 & 0.01 & 0.08 \\
Longitudinal & v\_start & -0.00 & 0.00 & 0.00 & 0.00 & 0.06 & \cellcolor{green!35}0.76 & 0.02 & 0.03 & 0.00 & 0.05 \\
Longitudinal & delta\_v & 0.01 & 0.01 & 0.00 & 0.01 & \cellcolor{green!35}0.51 & 0.00 & 0.15 & 0.01 & 0.02 & 0.00 \\
Longitudinal & a\_dur & 0.01 & 0.00 & -0.00 & 0.01 & \cellcolor{green!35}0.34 & 0.00 & 0.11 & 0.04 & 0.01 & -0.00 \\
Longitudinal & a\_pre & 0.01 & 0.02 & 0.00 & 0.01 & 0.02 & 0.00 & 0.01 & 0.12 & 0.00 & -0.00 \\
Longitudinal & a\_post & 0.00 & 0.00 & -0.00 & 0.00 & 0.06 & 0.00 & 0.03 & 0.11 & 0.00 & 0.01 \\
\hline
\end{tabular}

\end{table*}
The per-dimension results show that linearly encoded parameters tend 
to have a single \emph{dominant} latent dimension carrying most of 
their information, though never exclusively. The gap between the 
single-dimension $R^2$ and the full-latent linear $R^2$ for 
example, $y_\text{init}$ at $0.76$ versus $0.95$ indicates that 
the remaining variance is contributed by smaller signals distributed 
across other dimensions. The encoding is therefore primarily but not 
strictly axis-aligned and a few dimensions concentrate most of the 
information, with secondary contributions from others.

Two dimensions stand out as the primary carriers. 
Dimension $z_5$ jointly encodes $v_\text{start}$ and $v_\text{end}$ 
at nearly identical $R^2 \approx 0.75$, indicating that the model 
represents overall maneuver speed as a single property rather than 
encoding its endpoints separately. Dimension $z_0$ is multiplexed by 
construction, since it was assigned to scenario type during training 
(Section~\ref{app:model_details}); the per-dimension probe confirms that 
this dimension carries scenario-correlated quantities starting 
lane $y_\text{init}$ and lateral rate $\bar{\dot{y}}$ together, 
reflecting the deterministic coupling between maneuver type, 
direction, and starting lane in the training data. The target lane 
$y_\text{final}$ is carried by a separate dimension $z_9$, and the 
within-maneuver velocity change $\Delta v$ shares $z_4$ with the 
active-phase acceleration $a_\text{dur}$, again reflecting their 
physical coupling.

\begin{figure*}
    \centering
    \includegraphics[width=1\linewidth]{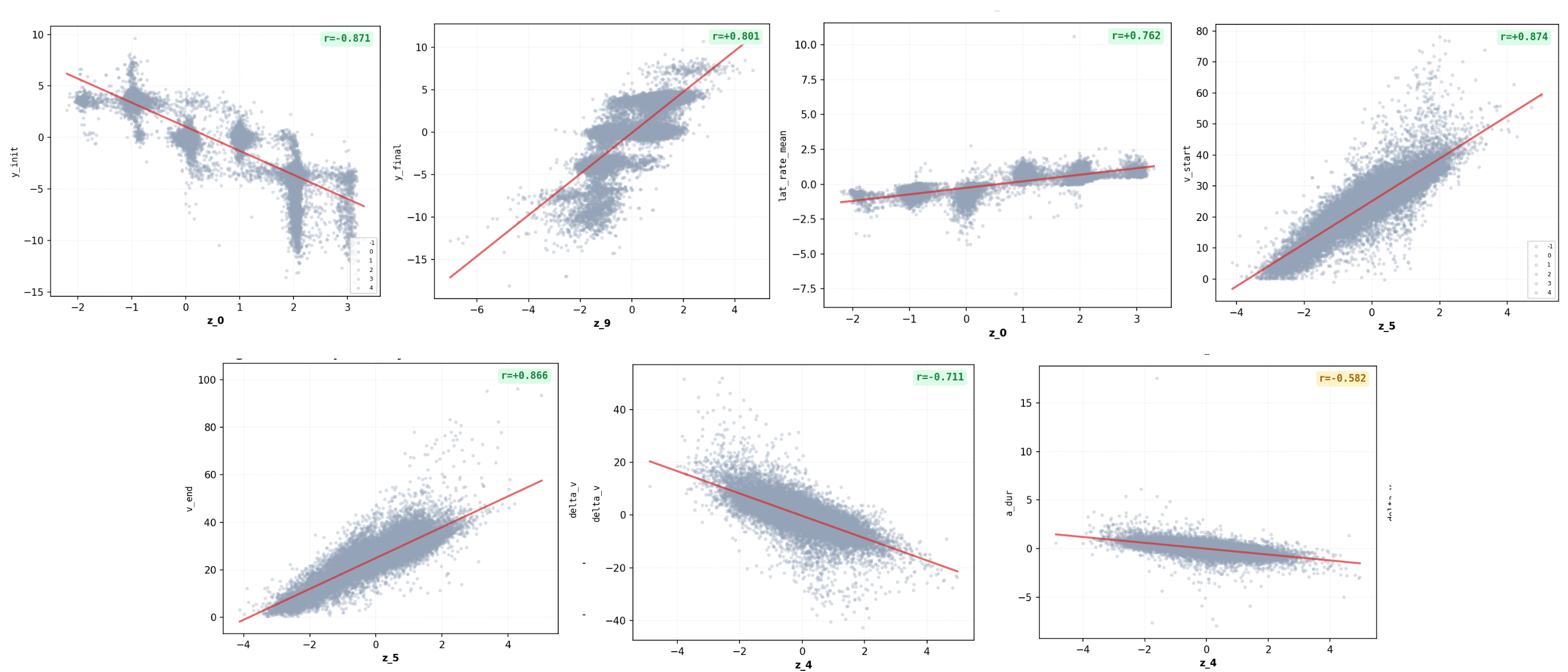}
    \caption{Per-sample relationships between dominant latent dimensions 
and physical parameters for the seven axis-aligned encodings.}
    \label{fig:axis_aligned_scatter}
\end{figure*}

Figure~\ref{fig:axis_aligned_scatter} visualizes the per-sample 
relationships underlying these correlation values. The high $R^2$ 
scores reflect genuinely tight relationships rather than wide clouds 
with strong slopes, and the figure additionally exposes a structural 
distinction not visible in the table: $z_0$ produces discrete cluster 
structure consistent with its supervised role as the scenario type 
dimension, while the remaining axis-aligned dimensions ($z_5$, $z_9$, 
$z_4$) produce continuous linear clouds, indicating that velocity, 
target lane, and velocity change are encoded uniformly across 
scenarios. This visualization confirms a dual encoding in the 
model: categorical for the supervised dimension and continuous for 
the unsupervised ones.
The non-linearly encoded parameters show a different picture. The 
pre- and post-maneuver accelerations peak only weakly on any single 
dimension ($z_7$, $R^2 \approx 0.11$), far below their full-latent 
nonlinear scores. The phase durations are the most extreme case: no 
single dimension reaches $R^2 = 0.06$. The model encodes timing 
information in a genuinely distributed and nonlinear form, with no 
axis-aligned component, consistent with the large probing gaps 
reported earlier.
\subsubsection{Layer 3}

\paragraph{Quantitative Cluster Scores}
Table~\ref{tab:layer_pca_silhouette} quantitatively confirms the trends previously observed in the visualization plots.

\begin{figure}[h]
    \centering
    \includegraphics[width=0.75\linewidth]{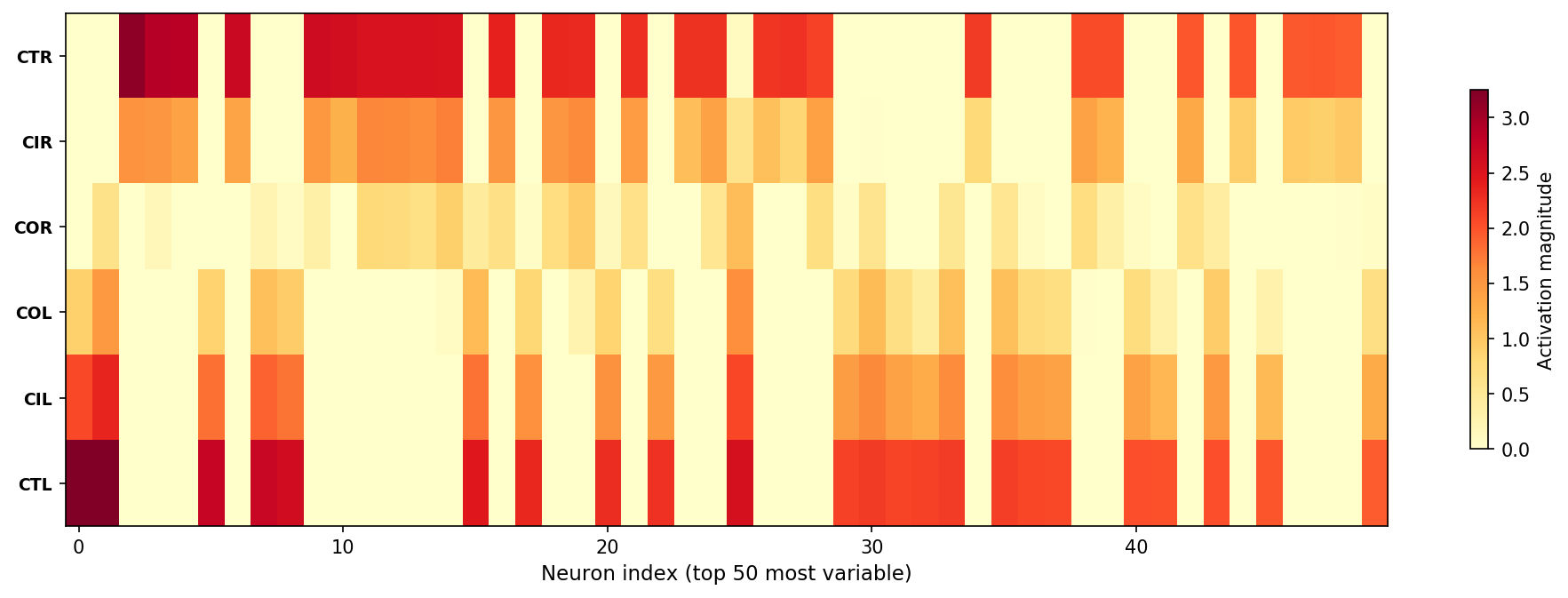}
    \caption{Activation Pattern in First Dense Layer.}
    \label{fig:circuit_heatmap}
\end{figure}

\begin{figure}[h]
    \centering
    \includegraphics[width=0.45\linewidth]{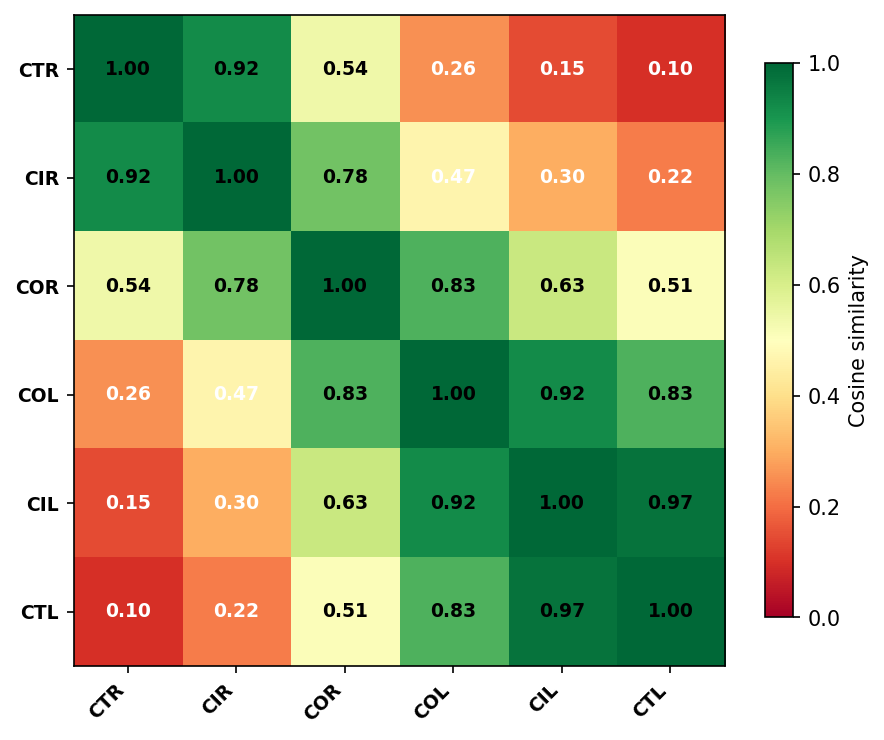}
    \caption{Cosine Similarity for Scenario Specific Activations.}
    \label{fig:circuit_similarity}
\end{figure}

\begin{table}[h]
\caption{Quantitative clustering measure using silhouette score and variance explained by the first principal component (PC1 EVR).}
\label{tab:layer_pca_silhouette}
\centering
\begin{tabular}{lcc}
\hline
\textbf{Layer} & \textbf{Silhouette Score} & \textbf{PC1 EVR (\%)} \\
\hline
Dense            & 0.557 & 22.68 \\
Reshape         & 0.557 & 22.68 \\
Convolution     & 0.413 & 23.57 \\
Convolution         & 0.339 & 26.90 \\
Convolution       & 0.517 & 24.36 \\
Flatten      & 0.517 & 24.36 \\
Dense                & 0.643 & 45.40 \\
Reshape               & 0.643 & 45.40 \\
\hline
\end{tabular}
\end{table}
\paragraph{Activation Pattern}
At the first dense layer ($n = 2500$), activation patterns exhibit clear 
sparse-coding structure. Approximately 30\% of neurons fire substantially 
for any given scenario, with the firing pattern being scenario-specific: 
distinct neuron subsets activate for distinct scenario types 
(Figure~\ref{fig:circuit_heatmap}). 
Further, the cosine similarity matrix between mean scenario activation vectors 
exhibits a ordered structure (Figure~\ref{fig:circuit_similarity}). 
Adjacent scenarios in the latent ordering  scenarios that differ in lateral 
extent but share direction have similarities above 0.9 (CTL 
vs CIL: 0.97; CTR vs CIRt: 0.92). 
Mirror-symmetric scenario pairs that differ in direction have similarities 
near zero (CTR vs CTL: 0.10; 
cIR vs CIL: 0.15).  
\paragraph{Trajectory Probing-Detailed}

\begin{figure}[h]
    \centering
    
    \begin{subfigure}{0.48\linewidth}
        \centering
        \includegraphics[width=\linewidth]{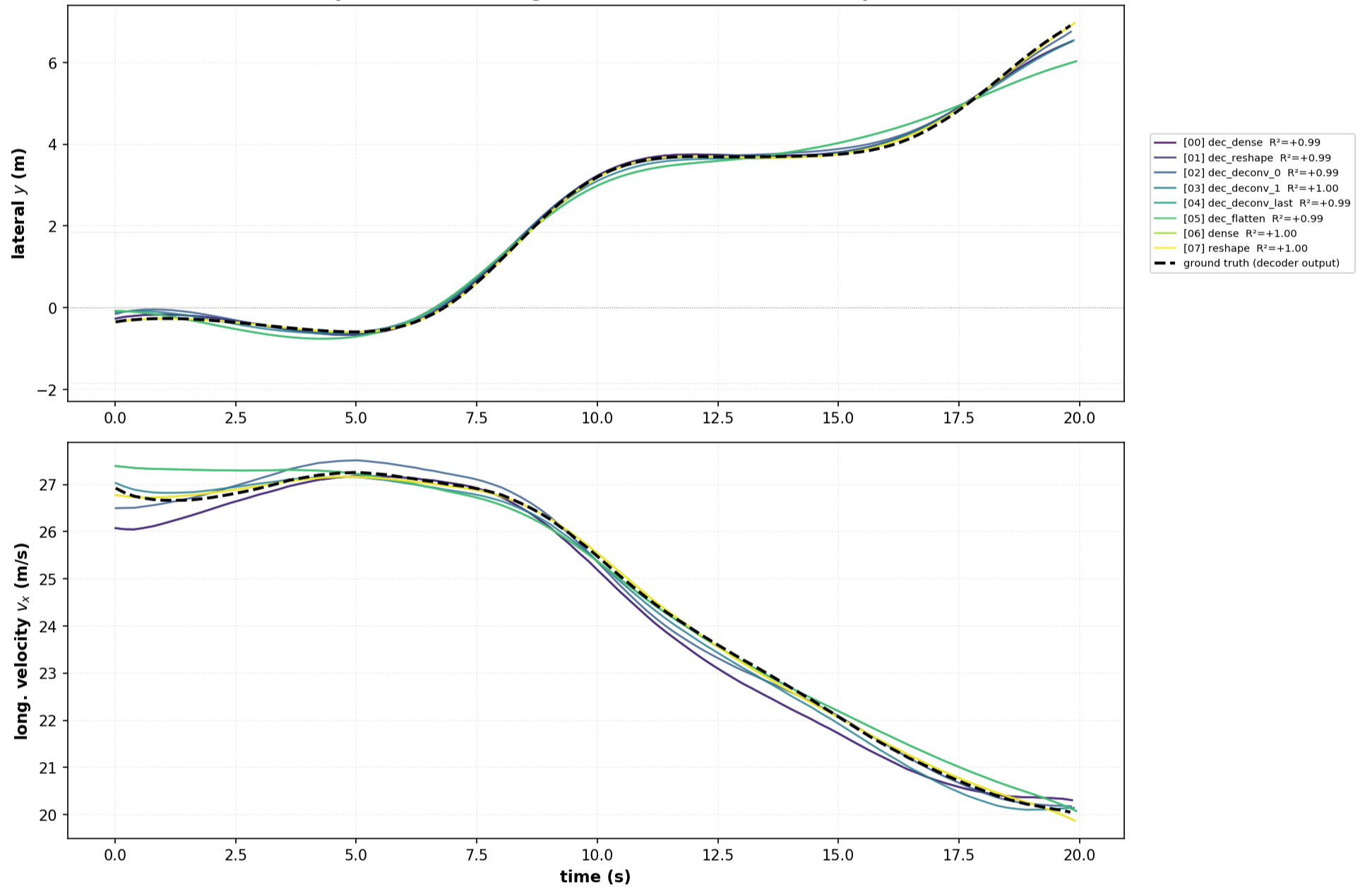}
        \caption{Cut-out left (COL)}
        \label{fig:per_scenario_cut_out_left}
    \end{subfigure}
    \hfill
    \begin{subfigure}{0.48\linewidth}
        \centering
        \includegraphics[width=\linewidth]{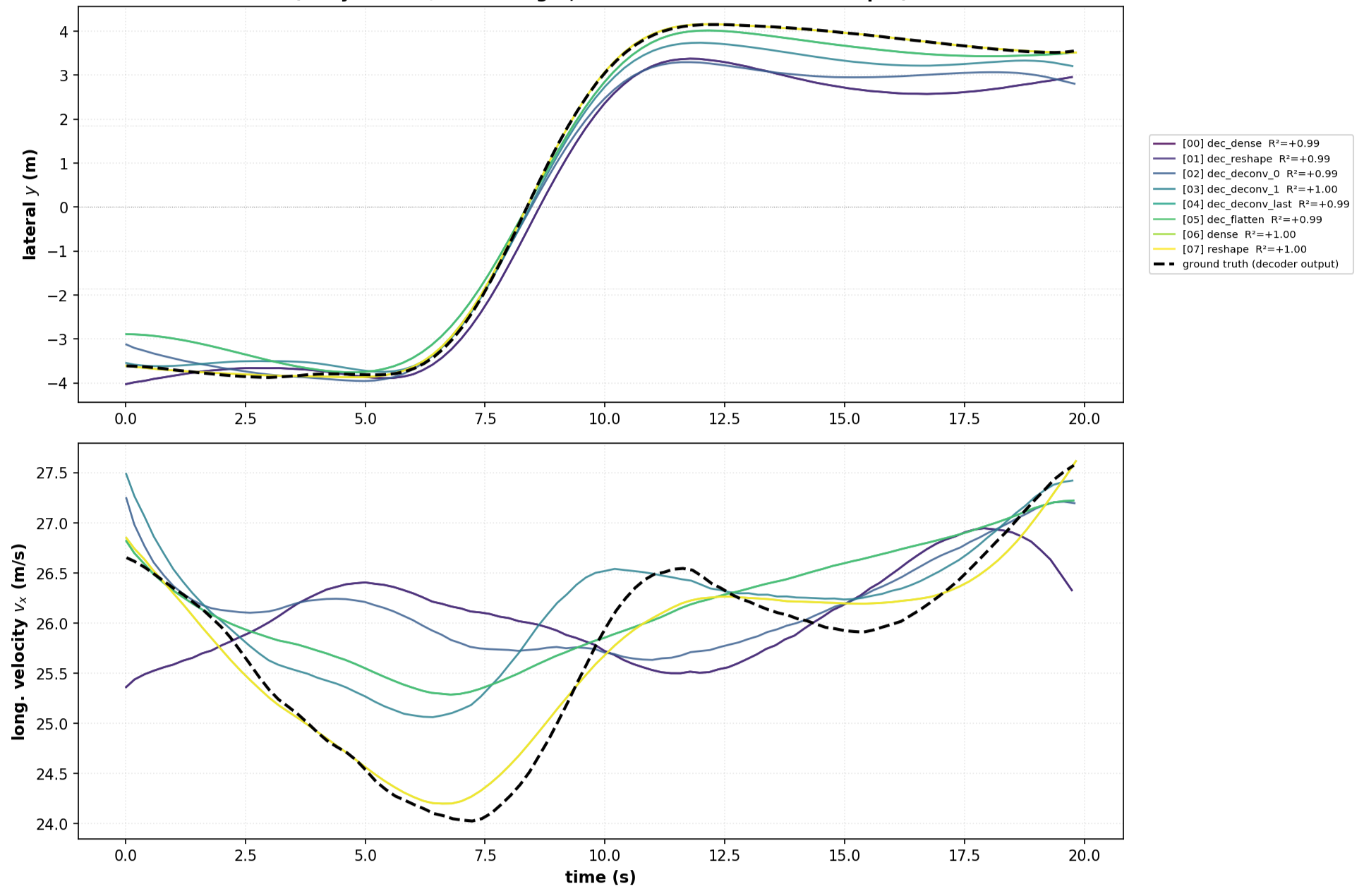}
        \caption{Cut-through left (CTL)}
        \label{fig:per_scenario_cut_through_left}
    \end{subfigure}
    
    \caption{Per-layer linear probe projections for representative scenarios.}
    \label{fig:per_scenario_combined}
\end{figure}

\begin{figure}[h]
    \centering
    \includegraphics[width=0.65\linewidth]{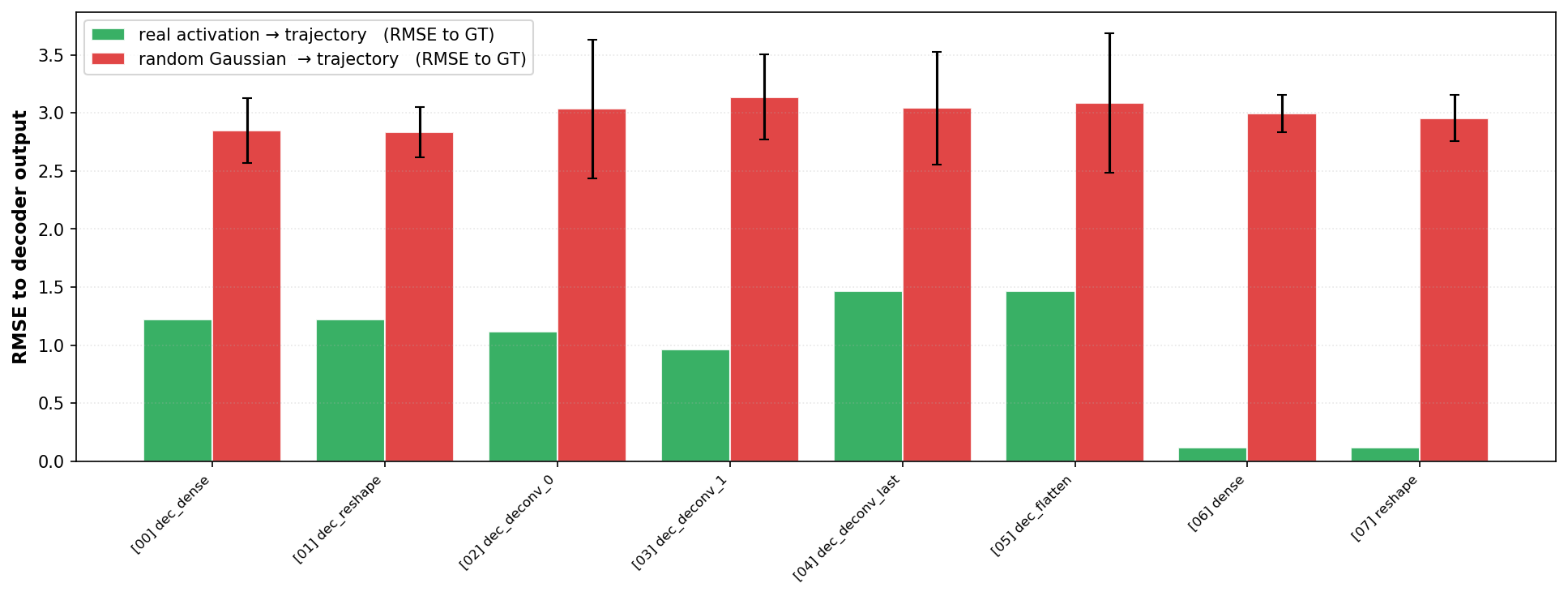}
    \caption{Sanity check for Layer-wise probes}
    \label{fig:sanity_check}
\end{figure}

Trajectory probing recovers the full output with $R^2 \in [0.99, 1.00]$ at every layer for every scenario, despite the per-parameter collapse in the middle convolutional layers. Trajectory information is therefore preserved at every layer even when its decomposition into individual parameters is not linearly accessible. A linear probe with $T \cdot C = 300$ output dimensions has substantially more degrees of freedom than per-parameter probes and can extract trajectory shape from any representation that preserves it. We illustrate the per layer behavior with two contrasting cases (Figures~\ref{fig:per_scenario_cut_out_left} and~\ref{fig:per_scenario_cut_through_left}). For COL, with a smoothly monotonic velocity profile, the projected trajectory at every layer including the first dense closely matches the decoder output ($\epsilon^{(l)}$ varies by less than $5\%$ across layers). For CTL, with coupled acceleration and deceleration phases, the velocity profile shows clear evolution: at early layers the projected velocity is smoother than the decoder output and misses the deceleration trough.The trough begins to form at intermediate convolutional layers and aligns with the output only at the final convolutional layer ($\epsilon^{(\text{dense})} > 3 \cdot \epsilon^{(\text{output})}$ on the velocity channel). The decoder thus appears to operate at different effective depths depending on output complexity.Simpler velocity profiles are recoverable from the first dense layer onward, while complex profiles require the convolutional layers to construct intermediate kinematic detail.
To validate that the layer-wise probes are signal-driven rather than producing input independent default outputs, we apply the same trained probes to random Gaussian vectors with matched per-feature mean and standard deviation; the resulting projections show roughly $4\times$ higher RMSE to the decoder output across all layers, ruling out the failure mode where high $R^2$ reflects probe mean prediction matching trajectory statistics by construction shown in Figure~\ref{fig:sanity_check}.

\subsection{Ablation Studies}
\subsubsection{Architectural Choice}
Here, we study the effect of changing the VAE regularization $\beta$ value. This changes how strongly the latent space is constrained. In theory, a $\beta$-VAE is expected to improve disentanglement by enforcing a more structured latent space. However, as shown in Table~\ref{tab:vae_beta_ablation}, the guided VAE gives more stable probing results. The $\beta$-VAE does not improve the parameter separation and gives lower probing scores for the available parameters. This suggests that a stronger regularization may remove useful information as well.
\begin{table}[h]
\centering
\caption{Parameter probing for different VAE .}
\label{tab:vae_beta_ablation}
\begin{tabular}{c c c c c}
\hline
 \textbf{Parameter} & \textbf{Autoencoder} & \textbf{Guided VAE} & \textbf{Standard VAE} & \textbf{$\beta$-VAE}  \\
        \hline
        $y_\text{init}$   & 0.44 & \fbox{0.762} & - & 0.32 \\
        $y_\text{final}$  & \fbox{0.74} & 0.637 & - & 0.66 \\
        $v_\text{start}$  & 0.45 & \fbox{0.760} & - & - \\
        $v_\text{end}$    & 0.50 & \fbox{0.743} & - & - \\
        $\Delta v$        & \fbox{0.57} & 0.501 & - & - \\
        $a_\text{dur}$    & \fbox{0.43} & 0.327 & - & - \\
        $\bar{\dot{y}}$   & 0.50 & \fbox{0.587} & - & 0.28 \\
\hline
\end{tabular}
\end{table}

\subsubsection{Latent Dimension}
For the latent dimension study, we compare a lower dimension of 6, the initial setting of 10, and a dimension of 12, which matches the number of parameters. As shown in Table~\ref{tab:vae_latent_dim_ablation}, increasing the latent dimension does not consistently improve the probing scores. Dimension 12 improves some individual parameters, especially the velocity terms, but performs worse for $\Delta v$ and $a_\text{dur}$. Dimension 6 also performs well for some parameters, but it is less balanced overall. Therefore, latent dimension 10 is kept as the final choice because it provides the most stable overall result.
\begin{table}[h]
\centering
\caption{Parameter probing comparison for different latent dimensions.}
\label{tab:vae_latent_dim_ablation}

\small
\renewcommand{\arraystretch}{1.2}

\begin{tabular}{|l|c|c|c|}
\hline
\textbf{Parameter} & \textbf{Latent Dim = 6} & \textbf{Latent Dim = 10} & \textbf{Latent Dim = 12} \\
\hline
$y_\text{init}$   & 0.75 & \fbox{0.762} & 0.75 \\
$y_\text{final}$  & 0.62 & 0.637 & \fbox{0.67} \\
$v_\text{start}$  & 0.54 & 0.760 & \fbox{0.79} \\
$v_\text{end}$    & 0.42 & 0.743 & \fbox{0.77} \\
$\Delta v$        & \fbox{0.69} & 0.501 & 0.33 \\
$a_\text{dur}$    & \fbox{0.43} & 0.327 & 0.26 \\
$\bar{\dot{y}}$   & \fbox{0.69} & 0.587 & 0.59 \\
\hline
\end{tabular}
\end{table}
\subsection{Cross-Architecture Demonstration}
\label{app:additional_implementations}
Here we show the results for the remaining architectures. The aim is to
show the protocol application to other model families, not to
benchmark them. We therefore use simple, minimal versions of each
architecture (interpretability analyses are typically
carried out on basic models). Since all layers have been discussed in
detail in the case studies above, we report only Layer 1,2 and 5 for the models. All models were trained on a workstation equipped with a 6-core CPU and an NVIDIA Quadro T2000 GPU (8 GB VRAM). The proposed protocol was executed on the same hardware.

The main change across
architectures is the choice of representation $z$ that the protocol
probes, summarised in Table~\ref{tab:z_defs}. This choice of $z$ is a design choice rather than a fixed rule because in most
architectures they allow several reasonable definitions, and the table is
meant as a guide. For the GAN, $z$ is drawn from the
prior.For the transformer we
include both the mean-pooled hidden state and the SAE-encoded features~\cite{} as representation
to see whether the SAE produces a sharper, more interpretable latent
than the raw pooled activation. The recurrent model is called
``world-model-inspired'' because it does not condition on actions and
is therefore not a world model in the strict sense; we include it as a
simple example of the sequential-latent case , so we can show
how the protocol adapts when the latent is a trajectory of hidden
states rather than a single vector.
\begin{table}[h]
\centering
\caption{Latent representation $z$ used as input to Layers 1--3 for each model.}
\small
\renewcommand{\arraystretch}{1.3}

\begin{tabular}{|l|l|}
\hline
\textbf{Model} & \textbf{Definition of $z$} \\
\hline
VAE                     & Latent vector $z$ \\
GAN                     & Sampled prior vector $z \sim \mathcal{N}(0, I)$ \\
Transformer (mean-pool) & Mean of final-layer residual stream over 100 positions \\
Transformer (SAE)       & SAE-encoded sparse features of the pooled hidden state \\
World model             & Final GRU hidden state $h_T$ \\
\hline
\end{tabular}

\label{tab:z_defs}
\end{table}

\subsubsection{Evaluation Results on GAN}
The architecture of the GAN used is given in Table~\ref{tab:gan}.

\begin{table}[h]
\centering
\caption{Network topology of the GAN model.}
\label{tab:gan}
\small
\renewcommand{\arraystretch}{1.3}

\begin{tabular}{|l|c|}
\hline
\multicolumn{2}{|c|}{\textbf{ Generator}} \\
\hline
\textbf{Layer} & \textbf{Output Shape} \\
\hline
Input                 & $10$ \\
Dense, ReLU           & $800$ \\
Reshape               & $25 \times 32$ \\
DeConv, ReLU          & $50 \times 64$ \\
DeConv, ReLU          & $100 \times 32$ \\
Conv, tanh            & $100 \times 3$ \\
\hline
\end{tabular}

\vspace{0.4cm}

\begin{tabular}{|l|c|}
\hline
\multicolumn{2}{|c|}{\textbf{Discriminator}} \\
\hline
\textbf{Layer} & \textbf{Output Shape} \\
\hline
Input                 & $100 \times 3$ \\
Conv, ReLU            & $50 \times 32$ \\
Conv, ReLU            & $25 \times 64$ \\
Flatten               & $1600$ \\
Dense, ReLU           & $32$ \\
Dense, sigmoid        & $1$ \\
\hline
\end{tabular}
\end{table}
\label{tab:gan_params}
\paragraph{Layer 1}
Table~\ref{tab:probe_gan} reports the results of Layer~1 probing on the
GAN. Linear probes recover the velocity-related parameters strongly
($v_\text{start}$, $v_\text{end}$, and $\Delta v$ all reach
$R^2_\text{lin} > 0.8$), while lateral and timing parameters are
considerably weaker, suggesting that the prior axes carry velocity
information almost linearly but encode lateral structure in a more
distributed form.

\begin{table}[h]
\centering
\setlength{\tabcolsep}{4pt}
\caption{Probing $R^2$ for physical parameters from $\mathbf{z}$ for GAN.}
\label{tab:probe_gan}

\begin{tabular}{l c c c c c}
\hline
\textbf{Parameter} & $R^2_\text{lin}$ & $R^2_\text{nl}$ & $\Delta R^2$ & $R^2_\text{null,lin}$ & $R^2_\text{null,nl}$ \\ \hline 
$y_\text{init}$         & 0.268 & -0.001 & 0.999 & -0.003 & +0.730 \\
$y_\text{final}$        & 0.429 & -0.001 & 0.999 & -0.002 & +0.570 \\
$t_\text{pre}$          & 0.112 & -0.001 & 0.807 & -0.006 & +0.695 \\
$t_\text{dur}$          & 0.033 & -0.001 & 0.740 & -0.007 & +0.707 \\
$t_\text{post}$         & 0.053 & -0.001 & 0.827 & -0.006 & +0.774 \\
$\text{lat\_rate}_\text{mean}$ & 0.547 & -0.001 & 0.887 & -0.013 & +0.339 \\
$v_\text{start}$        & \textbf{0.924} & -0.001 & 0.999 & -0.008 & +0.075 \\
$v_\text{end}$          & \textbf{0.903} & -0.001 & 0.998 & -0.010 & +0.095 \\
$\Delta v$              & \textbf{0.816} & -0.001 & 0.996 & -0.002 & +0.180 \\
$a_\text{pre}$          & 0.135 & -0.001 & 0.329 & -0.014 & +0.194 \\
$a_\text{dur}$          & 0.506 & -0.001 & 0.770 & -0.022 & +0.264 \\
$a_\text{post}$         & 0.110 & -0.001 & 0.333 & -0.012 & +0.223 \\
\hline
\end{tabular}
\end{table}
\paragraph{Layer 2}
Table~\ref{tab:best_latents_GAN} compares, for each parameter, the
single best latent dimension on the model against the
principal component of the data. Velocities are concentrated on a
single $z$ dimension to a comparable or greater degree than they are on
a principal component ($v_\text{start}$: $0.67$ vs.\ $0.61$), whereas
lateral and acceleration parameters are spread over multiple $z$ dimensions
and look weaker per-axis than PCA.
\begin{table}[htbp]
\centering
\caption{Best latent dimension contribution per feature for PCA and GAN.}
\begin{tabular}{lcc}
\hline
\textbf{Feature} & \textbf{PCA (best PC)} & \textbf{GAN (best $z$)} \\
\hline
$y_\text{init}$                 & 0.73 & 0.11 \\
$y_\text{final}$                & 0.64 & 0.14 \\
$v_\text{start}$                & 0.61 & 0.67 \\
$v_\text{end}$                  & 0.69 & 0.40 \\
$\Delta v$                      & 0.48 & 0.36 \\
$a_\text{dur}$                  & 0.28 & 0.14 \\
\hline
\end{tabular}
\label{tab:best_latents_GAN}
\end{table}
\paragraph{Layer 5}
The GAN's generated trajectories fail the physical-plausibility checks
across every scenario(shown in Table~\ref{tab:pvr_per_scenario_gan}), with PVR dropping by 69-89 percentage points
relative to the training distribution and ending in single digits for
all six classes. The collapse primarily due to lateral jerk
($p_{j_y}$ falls to 14--26\%), with longitudinal acceleration
($p_{a_x}$) and friction ($p_f$) also failing roughly half the time.
The lateral velocity check $p_{v_y}$ is the only metric the GAN largely
satisfies, indicating that the maneuvers happen at plausible lateral
speeds but are dynamically too abrupt.
\begin{table*}[h]
\centering
\caption{Per-scenario pass rates on training and generated trajectories of GAN.}
\label{tab:pvr_per_scenario_gan}

\footnotesize
\setlength{\tabcolsep}{3.2pt}
\renewcommand{\arraystretch}{1.05}

\begin{subtable}[t]{0.48\textwidth}
\centering
\begin{tabular}{l c c c c c c c}
\toprule
\textbf{Scen.} & \textbf{Src.} &
$p_f$ & $p_{a_y}$ & $p_{j_y}$ & $p_{a_x}$ & $p_{v_y}$ & \textbf{PVR} \\
\midrule
CIL & train    & 99.9 & 99.9 & 99.5 & 98.8 & 97.3 & 95.9 \\
    & gen      & 64.8 & 63.8 & 25.8 & 47.0 & 92.6 & 13.3 \\
    & $\Delta$ & -35.1 & -36.1 & -73.7 & -51.8 & -4.7 & \textbf{-82.6} \\
\addlinespace[2pt]

CIR & train    & 93.1 & 99.3 & 96.5 & 85.9 & 100.0 & 85.3 \\
    & gen      & 57.6 & 65.2 & 25.1 & 37.3 & 92.2 & 12.4 \\
    & $\Delta$ & -35.5 & -34.1 & -71.4 & -48.6 & -7.8 & \textbf{-72.9} \\
\addlinespace[2pt]

COL & train    & 99.9 & 99.9 & 99.8 & 99.7 & 99.9 & 99.5 \\
    & gen      & 58.7 & 57.9 & 21.8 & 40.9 & 93.3 & 10.1 \\
    & $\Delta$ & -41.3 & -42.1 & -78.1 & -58.8 & -6.6 & \textbf{-89.4} \\
\bottomrule
\end{tabular}
\end{subtable}
\hfill
\begin{subtable}[t]{0.48\textwidth}
\centering
\begin{tabular}{l c c c c c c c}
\toprule
\textbf{Scen.} & \textbf{Src.} &
$p_f$ & $p_{a_y}$ & $p_{j_y}$ & $p_{a_x}$ & $p_{v_y}$ & \textbf{PVR} \\
\midrule
COR & train    & 96.8 & 99.6 & 98.6 & 92.3 & 95.7 & 88.3 \\
    & gen      & 38.9 & 42.8 & 14.4 & 32.6 & 83.7 & 5.2 \\
    & $\Delta$ & -57.8 & -56.9 & -84.1 & -59.7 & -12.1 & \textbf{-83.0} \\
\addlinespace[2pt]

CTL & train    & 99.0 & 99.3 & 98.3 & 95.9 & 87.8 & 84.3 \\
    & gen      & 62.8 & 55.5 & 14.0 & 51.4 & 88.8 & 7.8 \\
    & $\Delta$ & -36.2 & -43.8 & -84.3 & -44.5 & +1.0 & \textbf{-76.5} \\
\addlinespace[2pt]

CTR & train    & 89.6 & 98.8 & 95.7 & 81.3 & 93.9 & 76.7 \\
    & gen      & 43.1 & 51.9 & 17.1 & 32.6 & 84.8 & 7.9 \\
    & $\Delta$ & -46.6 & -47.0 & -78.6 & -48.7 & -9.2 & \textbf{-68.7} \\
\bottomrule
\end{tabular}
\end{subtable}

\end{table*}
\subsubsection{Evaluation Results on Transformer-based}
The architecture of the basic decoder-based transformer is given in Table~\ref{tab:transformer_archi} and SAE used as choice for Probing in Table~\ref{tab:sae_archi} 

Note: The transformer and world-model baselines rely on input conditioning to generate trajectories. We initialize them using prefixes sampled from the held-out dataset. As expected, their outputs vary noticeably with the prefix length, since the amount of observed context directly constrains and shapes the subsequent predictions.
\begin{table}[h]
\centering
\caption{Network architecture of the transformer.}
\small
\renewcommand{\arraystretch}{1.3}
\begin{tabular}{|l|c|}
\hline
\textbf{Layer} & \textbf{Output Shape} \\
\hline
Input                              & $100 \times 3$ \\
Linear (token embedding)           & $100 \times 64$ \\
+ Positional embedding             & $100 \times 64$ \\
Self-Attention (4 heads), causal   & $100 \times 64$ \\
Feed-Forward MLP (64 $\to$ 128 $\to$ 64), GELU & $100 \times 64$ \\
Self-Attention (4 heads), causal   & $100 \times 64$ \\
Feed-Forward MLP (64 $\to$ 128 $\to$ 64), GELU & $100 \times 64$ \\
LayerNorm                          & $100 \times 64$ \\
Linear (output head)               & $100 \times 3$ \\
\hline
\end{tabular}
\label{tab:transformer_archi}
\end{table}

\begin{table}[h]
\centering
\caption{Network architecture of the sparse autoencoder trained on the
transformer's mean-pooled hidden state.}
\small
\renewcommand{\arraystretch}{1.3}
\begin{tabular}{|l|c|l|c|}
\hline
\textbf{Encoder Layer} & \textbf{Output Shape} &
\textbf{Decoder Layer} & \textbf{Output Shape} \\
\hline
Input (pooled hidden)       & 64  & Input (sparse code)        & 64  \\
Subtract $b_\mathrm{dec}$   & 64  & Linear ($W_\mathrm{dec}$)  & 64  \\
Linear ($W_\mathrm{enc}$)   & 64  & Add $b_\mathrm{dec}$       & 64  \\
Add $b_\mathrm{enc}$, ReLU  & 64  & ---                        & --- \\
\hline
\end{tabular}
\label{tab:sae_archi}
\end{table}

\paragraph{Layer 1}
The results from Table~\ref{tab:tf_l1_probing} shows that mean-pooled hidden state already captures most of the
trajectory-defining information in a close-to linear way: $y_\text{init}$,
$v_\text{start}$, $v_\text{end}$, and $\Delta v$ all achieve higher
$R^2$, while timing and acceleration variables remain only weakly
recoverable. Adding the SAE here only gives representations of similar
quality, with no clear improvement on the strongly encoded variables and
a slight drop on a few of them.
\begin{table}[h]
\centering
\caption{Probing results ($R^2$) for Transformer representations at Layer 1.}
\label{tab:tf_l1_probing}

\small
\setlength{\tabcolsep}{5pt}
\renewcommand{\arraystretch}{1.15}

\begin{tabular}{lcc}
\hline
\textbf{Parameter} & \textbf{TF mean-pool ($d=64$)} & \textbf{TF SAE ($d=64$)} \\
\hline
$y_\text{init}$         & \textbf{0.956} & 0.934 \\
$y_\text{final}$        & 0.793 & \textbf{0.805} \\
$t_\text{pre}$          & 0.205 & \textbf{0.206} \\
$t_\text{dur}$          & \textbf{0.267} & 0.251 \\
$t_\text{post}$         & \textbf{0.452} & 0.435 \\
$v_\text{start}$        & \textbf{0.966} & 0.958 \\
$v_\text{end}$          & \textbf{0.935} & 0.934 \\
$a_\text{pre}$          & \textbf{0.276} & 0.252 \\
$a_\text{dur}$          & \textbf{0.109} & 0.093 \\
$a_\text{post}$         & 0.063 & \textbf{0.064} \\
$\text{lat\_rate}_\text{mean}$ & \textbf{-0.022} & -0.044 \\
$\Delta v$              & \textbf{0.873} & 0.871 \\
\hline
\end{tabular}

\end{table}

\paragraph{Layer 2}
From Table~\ref{tab:l2_tf_meanpool} and Table~\ref{tab:l2_tf_sae} shows that for variables that the transformer already encodes strongly
($y_\text{init}$, $y_\text{final}$, velocities, $\Delta v$), PCA on the
raw representation performs similarly or slightly better.
\begin{table}[htbp]
\centering
\caption{Layer 2 probing results for TF mean-pool representation.}
\label{tab:l2_tf_meanpool}

\small
\setlength{\tabcolsep}{5pt}
\renewcommand{\arraystretch}{1.15}

\begin{tabular}{lccc}
\hline
\textbf{Parameter} & $R^2_\text{model}$ & $R^2_\text{pca}$ & $\Delta$ \\
\hline
$y_\text{init}$        & 0.956 & 0.999 & -0.043 \\
$y_\text{final}$       & 0.793 & 0.999 & -0.207 \\
$t_\text{pre}$         & 0.205 & 0.184 & 0.020 \\
$t_\text{dur}$         & 0.267 & 0.127 & 0.140 \\
$t_\text{post}$        & 0.452 & 0.303 & 0.148 \\
$v_\text{start}$       & 0.966 & 1.000 & -0.034 \\
$v_\text{end}$         & 0.935 & 1.000 & -0.065 \\
$a_\text{pre}$         & 0.276 & 0.425 & -0.148 \\
$a_\text{dur}$         & 0.109 & 0.052 & 0.057 \\
$a_\text{post}$        & 0.063 & 0.029 & 0.034 \\
$\text{lat\_rate}_\text{mean}$ & -0.022 & -0.027 & 0.005 \\
$\Delta v$             & 0.873 & 1.000 & -0.127 \\
\hline
\end{tabular}
\end{table}

\begin{table}[htbp]
\centering
\caption{Layer 2 probing results for TF SAE representation.}
\label{tab:l2_tf_sae}

\small
\setlength{\tabcolsep}{5pt}
\renewcommand{\arraystretch}{1.15}

\begin{tabular}{lccc}
\hline
\textbf{Parameter} & $R^2_\text{model}$ & $R^2_\text{pca}$ & $\Delta$ \\
\hline
$y_\text{init}$        & 0.934 & 0.999 & -0.065 \\
$y_\text{final}$       & 0.805 & 0.999 & -0.194 \\
$t_\text{pre}$         & 0.206 & 0.184 & 0.021 \\
$t_\text{dur}$         & 0.251 & 0.127 & 0.124 \\
$t_\text{post}$        & 0.435 & 0.303 & 0.131 \\
$v_\text{start}$       & 0.958 & 1.000 & -0.042 \\
$v_\text{end}$         & 0.934 & 1.000 & -0.066 \\
$a_\text{pre}$         & 0.252 & 0.425 & -0.172 \\
$a_\text{dur}$         & 0.093 & 0.052 & 0.041 \\
$a_\text{post}$        & 0.064 & 0.029 & 0.035 \\
$\text{lat\_rate}_\text{mean}$ & -0.044 & -0.027 & -0.017 \\
$\Delta v$             & 0.871 & 1.000 & -0.129 \\
\hline
\end{tabular}
\end{table}
\paragraph{Layer 5}
Generated trajectories show large physical-plausibility drops in every
scenario Table~\ref{tab:pvr_per_scenario_tf},indicating that the autoregressive
predictions are dynamically inconsistent failing in multiple axes. This may be because of the simplicity of architecture used . We also found depending on the prefix length, it produces better results.

\textbf{Note on model quality}: The transformer and SAE used here are simple, lightly-trained
architectures intended to demonstrate the protocol's applicability to
this case, not strong baselines. The numerical values should therefore
be read as illustrative rather than as benchmarks: a more carefully
trained model would be expected to score higher on Layers 1, 2, and 5,
but the questions the protocol asks whether the representation
encodes physical structure, better than a PCA baseline, and produces
plausible outputs would be answered the same way and in the same
form. Similar case applies to the next model.

\begin{table*}[h]
\centering
\caption{Per-scenario pass rates on training and generated trajectories of Transformer-based models.}
\label{tab:pvr_per_scenario_tf}
\footnotesize
\setlength{\tabcolsep}{2.8pt}
\renewcommand{\arraystretch}{1.05}
\begin{minipage}[t]{0.49\textwidth}
\centering
\begin{tabular}{l c c c c c c c}
\toprule
\textbf{Scen.} & \textbf{Src.} &
$p_f$ & $p_{a_y}$ & $p_{j_y}$ & $p_{a_x}$ & $p_{v_y}$ & \textbf{PVR} \\
\midrule
CIL & train & 99.9 & 99.8 & 99.5 & 98.8 & 97.3 & 96.0 \\
    & gen   & 73.8 & 73.3 & 72.1 & 19.1 & 72.9 & 16.6 \\
    & $\Delta$ & -26.0 & -26.6 & -27.4 & -79.7 & -24.4 & \textbf{-79.3} \\
\addlinespace[2pt]
CIR & train & 93.0 & 99.2 & 96.5 & 86.0 & 100.0 & 85.3 \\
    & gen   & 85.9 & 90.0 & 88.5 & 33.3 & 90.3 & 29.6 \\
    & $\Delta$ & -7.1 & -9.2 & -8.1 & -52.6 & -9.7 & \textbf{-55.8} \\
\addlinespace[2pt]
COL & train & 99.9 & 99.9 & 99.8 & 99.7 & 100.0 & 99.5 \\
    & gen   & 91.5 & 91.1 & 90.3 & 73.8 & 91.0 & 73.2 \\
    & $\Delta$ & -8.4 & -8.8 & -9.5 & -25.9 & -8.9 & \textbf{-26.3} \\
\bottomrule
\end{tabular}
\end{minipage}%
\hfill
\begin{minipage}[t]{0.49\textwidth}
\centering
\begin{tabular}{l c c c c c c c}
\toprule
\textbf{Scen.} & \textbf{Src.} &
$p_f$ & $p_{a_y}$ & $p_{j_y}$ & $p_{a_x}$ & $p_{v_y}$ & \textbf{PVR} \\
\midrule
COR & train & 97.0 & 99.7 & 98.6 & 92.4 & 95.9 & 88.6 \\
    & gen   & 44.2 & 54.6 & 39.2 & 7.3 & 72.4 & 3.8 \\
    & $\Delta$ & -52.8 & -45.0 & -59.4 & -85.1 & -23.5 & \textbf{-84.8} \\
\addlinespace[2pt]
CTL & train & 99.1 & 99.1 & 98.0 & 96.0 & 87.5 & 84.3 \\
    & gen   & 69.3 & 68.8 & 66.8 & 53.2 & 64.1 & 47.7 \\
    & $\Delta$ & -29.8 & -30.3 & -31.2 & -42.7 & -23.4 & \textbf{-36.5} \\
\addlinespace[2pt]
CTR & train & 88.6 & 98.8 & 95.2 & 79.7 & 93.4 & 74.9 \\
    & gen   & 59.8 & 67.2 & 56.4 & 9.9 & 76.9 & 7.5 \\
    & $\Delta$ & -28.9 & -31.6 & -38.8 & -69.8 & -16.5 & \textbf{-67.4} \\
\bottomrule
\end{tabular}
\end{minipage}
\end{table*}
\subsubsection{Evaluation Results on World-Model-Inspired GRU Rollout}
Architecture of demonstration model given in Table~\ref{tab:wm_archi}.
\begin{table}[h]
\centering
\caption{Network architecture of the recurrent world model.}
\small
\renewcommand{\arraystretch}{1.3}
\begin{tabular}{|l|c|}
\hline
\textbf{Layer} & \textbf{Output Shape} \\
\hline
Input                       & $100 \times 3$ \\
Linear, GELU (encoder)      & $100 \times 64$ \\
Linear (encoder)            & $100 \times 32$ \\
GRU cell (recurrent dynamics) & $100 \times 32$ \\
Linear, GELU (decoder)      & $100 \times 64$ \\
Linear (decoder)            & $100 \times 3$ \\
\hline
\end{tabular}
\label{tab:wm_archi}
\end{table}
\paragraph{Layer 1}
For Layer 1 we also show the results of hidden state at different steps as shown in 
Table~\ref{tab:wm_l1_probing}. Here, the probe results can be interpreted as showing which part of the rollout encodes each physical property. Variables associated with the beginning of the trajectory ($y_\text{init}$, $v_\text{start}$, $a_\text{pre}$) are best recovered from $h_{t10}$, and their decodability decreases steadily toward $h_{t99}$ as the GRU gradually forgets its initial state. In contrast, end-of-trajectory variables ($y_\text{final}$, $v_\text{end}$, $\Delta v$) follow the opposite trend, becoming most accurately decoded at $h_{t99}$.  Quantities linked to the middle of the rollout ($t_\text{pre}$, $t_\text{dur}$, $a_\text{dur}$) peak around $h_{t50}$.

\begin{table}[h]
\centering
\caption{Probing results ($R^2$) for World Model hidden state at different time steps (Layer 1, $d=32$).}
\label{tab:wm_l1_probing}

\small
\setlength{\tabcolsep}{4.5pt}
\renewcommand{\arraystretch}{1.15}

\begin{tabular}{lccc}
\hline
\textbf{Parameter} & \textbf{$h_{t10}$} & \textbf{$h_{t50}$} & \textbf{$h_{t99}$} \\
\hline
$y_\text{init}$         & \textbf{0.982} & 0.571 & 0.131 \\
$y_\text{final}$        & 0.475 & 0.779 & \textbf{1.000} \\
$t_\text{pre}$          & 0.261 & \textbf{0.348} & 0.171 \\
$t_\text{dur}$          & 0.146 & \textbf{0.232} & 0.187 \\
$t_\text{post}$         & 0.515 & \textbf{0.666} & 0.638 \\
$v_\text{start}$        & \textbf{0.984} & 0.837 & 0.554 \\
$v_\text{end}$          & 0.793 & 0.960 & \textbf{1.000} \\
$a_\text{pre}$          & \textbf{0.409} & 0.146 & 0.047 \\
$a_\text{dur}$          & 0.246 & \textbf{0.401} & 0.016 \\
$a_\text{post}$         & \textbf{0.482} & 0.489 & 0.274 \\
$\text{lat\_rate}_\text{mean}$ & \textbf{0.008} & 0.013 & 0.006 \\
$\Delta v$              & 0.699 & 0.924 & \textbf{0.948} \\
\hline
\end{tabular}

\end{table}

\paragraph{Layer 2}
The world model's hidden state largely matches PCA on the strongly
encoded parameters as shown in Table~\ref{tab:l2_wm_t50} and Table~\ref{tab:l2_wm_t99} at $h_{t99}$ ($y_\text{final}$, $v_\text{end}$,
$\Delta v$ with $\Delta \approx 0$) but underperforms PCA on the
beginning-of-trajectory properties at the same timestep, which is
expected because step 99 the GRU has overwritten most of the early
state, while PCA retains it because it sees the full flattened
trajectory. 
\begin{table}[h]
\centering
\caption{Layer 2 probing results for World Model ($t_{50}$).}
\label{tab:l2_wm_t50}

\small
\setlength{\tabcolsep}{5pt}
\renewcommand{\arraystretch}{1.15}

\begin{tabular}{lccc}
\hline
\textbf{Parameter} & $R^2_\text{model}$ & $R^2_\text{pca}$ & $\Delta$ \\
\hline
$y_\text{init}$        & 0.571 & 0.996 & -0.424 \\
$y_\text{final}$       & 0.779 & 0.999 & -0.220 \\
$t_\text{pre}$         & 0.348 & 0.342 & 0.006 \\
$t_\text{dur}$         & 0.232 & 0.234 & -0.002 \\
$t_\text{post}$        & 0.666 & 0.668 & -0.001 \\
$v_\text{start}$       & 0.837 & 1.000 & -0.162 \\
$v_\text{end}$         & 0.960 & 1.000 & -0.040 \\
$a_\text{pre}$         & 0.146 & 0.549 & -0.403 \\
$a_\text{dur}$         & 0.401 & 0.400 & 0.001 \\
$a_\text{post}$        & 0.489 & 0.536 & -0.047 \\
$\text{lat\_rate}_\text{mean}$ & 0.013 & 0.013 & 0.000 \\
$\Delta v$             & 0.924 & 1.000 & -0.076 \\
\hline
\end{tabular}
\end{table}

\begin{table}[h]
\centering
\caption{Layer 2 probing results for World Model ($t_{99}$).}
\label{tab:l2_wm_t99}

\small
\setlength{\tabcolsep}{5pt}
\renewcommand{\arraystretch}{1.15}

\begin{tabular}{lccc}
\hline
\textbf{Parameter} & $R^2_\text{model}$ & $R^2_\text{pca}$ & $\Delta$ \\
\hline
$y_\text{init}$        & 0.131 & 0.996 & -0.865 \\
$y_\text{final}$       & 1.000 & 0.999 & 0.001 \\
$t_\text{pre}$         & 0.171 & 0.342 & -0.171 \\
$t_\text{dur}$         & 0.187 & 0.234 & -0.047 \\
$t_\text{post}$        & 0.638 & 0.668 & -0.030 \\
$v_\text{start}$       & 0.554 & 1.000 & -0.445 \\
$v_\text{end}$         & 1.000 & 1.000 & 0.000 \\
$a_\text{pre}$         & 0.047 & 0.549 & -0.501 \\
$a_\text{dur}$         & 0.016 & 0.400 & -0.384 \\
$a_\text{post}$        & 0.274 & 0.536 & -0.261 \\
$\text{lat\_rate}_\text{mean}$ & 0.006 & 0.013 & -0.007 \\
$\Delta v$             & 0.948 & 1.000 & -0.052 \\
\hline
\end{tabular}
\end{table}
\paragraph{Layer 5}
Here also the model shows low physical Plausibility as shown in Table~\ref{tab:pvr_per_scenario_wm}

\begin{table*}[h]
\centering
\caption{Per-scenario pass rates on training and generated trajectories of World-Model-Inspired}
\label{tab:pvr_per_scenario_wm}

\footnotesize
\setlength{\tabcolsep}{2.8pt}
\renewcommand{\arraystretch}{1.05}

\begin{subtable}[t]{0.49\textwidth}
\centering
\begin{tabular}{l c c c c c c c}
\toprule
\textbf{Scen.} & \textbf{Src.} &
$p_f$ & $p_{a_y}$ & $p_{j_y}$ & $p_{a_x}$ & $p_{v_y}$ & \textbf{PVR} \\
\midrule

CIL & train & 99.9 & 99.8 & 99.5 & 98.8 & 97.3 & 96.0 \\
    & gen   & 80.9 & 80.3 & 79.2 & 24.3 & 74.0 & 22.5 \\
    & $\Delta$ & -18.9 & -19.5 & -20.3 & -74.5 & -23.3 & \textbf{-73.4} \\
\addlinespace[2pt]

CIR & train & 93.0 & 99.2 & 96.5 & 86.0 & 100.0 & 85.3 \\
    & gen   & 15.0 & 15.8 & 13.2 & 14.5 & 14.5 & 12.8 \\
    & $\Delta$ & -78.1 & -83.4 & -83.3 & -71.4 & -85.5 & \textbf{-72.5} \\
\addlinespace[2pt]

COL & train & 99.9 & 99.9 & 99.8 & 99.7 & 100.0 & 99.5 \\
    & gen   & 49.2 & 49.3 & 46.7 & 43.7 & 55.3 & 42.9 \\
    & $\Delta$ & -50.7 & -50.7 & -53.1 & -56.0 & -44.6 & \textbf{-56.6} \\
\bottomrule
\end{tabular}
\end{subtable}
\hfill
\begin{subtable}[t]{0.49\textwidth}
\centering
\begin{tabular}{l c c c c c c c}
\toprule
\textbf{Scen.} & \textbf{Src.} &
$p_f$ & $p_{a_y}$ & $p_{j_y}$ & $p_{a_x}$ & $p_{v_y}$ & \textbf{PVR} \\
\midrule

COR & train & 97.0 & 99.7 & 98.6 & 92.4 & 95.9 & 88.6 \\
    & gen   & 3.4 & 3.5 & 3.0 & 2.6 & 3.4 & 2.0 \\
    & $\Delta$ & -93.6 & -96.2 & -95.6 & -89.8 & -92.4 & \textbf{-86.6} \\
\addlinespace[2pt]

CTL & train & 99.1 & 99.1 & 98.0 & 96.0 & 87.5 & 84.3 \\
    & gen   & 78.7 & 78.5 & 74.1 & 61.9 & 75.6 & 55.8 \\
    & $\Delta$ & -20.4 & -20.7 & -23.8 & -34.1 & -11.8 & \textbf{-28.5} \\
\addlinespace[2pt]

CTR & train & 88.6 & 98.8 & 95.2 & 79.7 & 93.4 & 74.9 \\
    & gen   & 1.3 & 1.3 & 1.0 & 1.1 & 1.2 & 0.6 \\
    & $\Delta$ & -87.4 & -97.5 & -94.2 & -78.5 & -92.3 & \textbf{-74.3} \\
\bottomrule
\end{tabular}
\end{subtable}

\end{table*}
\subsection{Comparison of WOSAC and Waymo E2E benchmarks with the proposed five-layer evaluation protocol}\label{sec:sota_comparison}
The comparison between the existing WOSAC/Waymo E2E evaluation 
framework and our proposed five-layer protocol is summarized in 
Table~\ref{tab:evaluation_protocol}.
\begin{table}[]
\centering
\caption{Comparison of WOSAC and Waymo E2E benchmarks with the proposed five-layer evaluation protocol.}

\label{tab:evaluation_protocol}

\resizebox{\columnwidth}{!}{%
\begin{tabular}{
    p{3.2cm}
    p{5.0cm}
    p{6.0cm}
    p{5.0cm}
}
\toprule
\textbf{Evaluation Dimension} &
\textbf{WOSAC \& Waymo E2E (SOTA Benchmarks)} &
\textbf{Proposed 5-Layer Protocol (Explicit \& Autoregressive)} &
\textbf{Key Advantage} \\
\midrule

\textbf{Target Scope} &
\textbf{Output Realism:} Evaluates final trajectories using aggregated metrics
(Realism Meta-Metric, Rater Feedback Score). &
\textbf{Internal Mechanics \& Physical Bounds:} Evaluates internal
representations alongside output constraints. &
Reveals \emph{how} the model generates scenarios, not just \emph{what} it outputs. \\

\midrule

\textbf{Applicability to Autoregressive Models} &
Evaluates final trajectory outputs directly. &
Uses internal hidden states (e.g., mean-pooled Transformer activations or
RNN step hidden states) as a representative latent vector. &
Extends internal interpretability to models without an explicit latent variable. \\

\midrule

\textbf{Layer 1: Kinematic Alignment} &
Not measured internally. &
Tests if physical parameters (speed, lateral rate, endpoints) are linearly
or nonlinearly encoded in hidden states. &
Proves whether the network explicitly learns physical concepts. \\

\midrule

\textbf{Layer 2: Statistical Baseline} &
Not measured. &
Compares internal representations against standard PCA data compression. &
Verifies the model learns meaningful driving structure rather than
mirroring dataset bias. \\

\midrule

\textbf{Layer 3: Feature Emergence} &
Not measured. &
Tracks feature formation across network layers (Transformers) or time
steps (RNNs). &
Identifies where in the network scenario structures and maneuvers form. \\

\midrule

\textbf{Layer 4: Latent / State Steering} &
Not measured. &
Applies systematic nudges to hidden states to observe changes in generated outputs. &
Tests whether the model can be reliably controlled toward target maneuvers. \\

\midrule

\textbf{Layer 5: Physical Plausibility} &
Evaluates speed/acceleration distributions. &
Checks strict, deterministic vehicle dynamics limits
(friction bounds, maximum lateral acceleration, maximum jerk). &
Exposes physically impossible trajectories that look visually realistic
in displacement metrics. \\

\bottomrule
\end{tabular}%
}

\end{table}
Additionally, we computed both the \textbf{WOSAC kinematic NLL} and \textbf{Layer 5 PVR} on the same generated trajectories for all four architectures on ExiD. We use the WOSAC kinematic subset (distributional NLL over speed, acceleration, jerk, and lateral displacement) but exclude interaction/map-based metrics (collision, off-road, TTC).

\begin{table}[]
\centering
\caption{Comparison of WOSAC and PVR.}
\label{tab:wosac_pvr}

\begin{tabular}{|l|c|c|}
\hline
\textbf{Model} & \textbf{WOSAC $\downarrow$} & \textbf{$\Delta$PVR $\uparrow$} \\ \hline
\textbf{VAE} & \textbf{2.9} & \textbf{+10.9\%} \\ \hline
GAN & 7.7 & -85.6\% \\ \hline
Transformer & 9.2 & -85.9\% \\ \hline
WorldModel & 9.5 & -64.9\% \\ \hline
\end{tabular}

\end{table}

\textbf{Key observations}

\begin{itemize}

    \item \textbf{Consistent ranking.} Both metrics identify the \textbf{VAE} as the best-performing model, supporting that PVR captures a meaningful quality signal rather than an artefact of threshold selection.

    \item \textbf{PVR reveals per-trajectory failures.} Although the GAN achieves lower WOSAC NLL than the Transformer and WorldModel, all three have near-zero PVR, indicating that many individual trajectories violate physical constraints despite matching the overall data distribution.

    \item \textbf{Safety interpretation.} PVR measures the proportion of generated trajectories that satisfy physically executable vehicle dynamics, providing an interpretable percentage. In contrast, WOSAC NLL is a relative distributional score that requires comparison with other models.

    \item \textbf{Complementary metrics.} We recommend using both metrics: \textbf{WOSAC} for distributional realism and \textbf{PVR} for interpretable, safety-oriented validation and diagnosis.

\end{itemize}

\subsection{How to Apply the Protocol to a Model}

\paragraph{Objective:}The protocol's main goal is to understand what the model has learned about the underlying driving behavior, whether
kinematic features are represented inside the model and if so where they appear, and whether they can be used to understand or influence generation.

The section now describes how to apply the protocol to a scenario generation model.

\subsubsection{Prerequisites}

Three things are required before any layer can be run:

\begin{itemize}
    \item A trained generative model with  access to its forward paths.
    \item A held-out dataset of real trajectories in the same domain the model was trained on. 
    \item A physical-parameter extractor that maps data to interpretable kinematic descriptors. The protocol is independent of the choice of extractor and will work with any expert model.
\end{itemize} 

\subsubsection{Procedure}
\begin{enumerate}
    \item \textbf{Define $z$ and $g$.}
    State which forward path produces $z$, its dimension $d$, and which
    path constitutes $g$. For architectures without a single explicit
    latent, define a representative vector, e.g.,
    $z=h_T$ for recurrent models or
    $z=\mathrm{mean}_t H^L$ for transformers, and report at least one
    robustness alternative.

    \item \textbf{Build the analysis matrices.}
    Encode the held-out dataset to obtain
    $Z$ and apply the physical-parameter
    extractor to obtain $P\in\mathbb{R}^{N\times M}$.
    For decoder-only latent models, sample $z$ from the prior, decode
    through $g$, and extract $P$ from the generated trajectories.

    \item \textbf{Run Layers 1-5.}
    Apply the five layers shown in Table~\ref{tab:layer_summary}, adapting only the representation and
    intervention mechanism to the model's architecture.
\end{enumerate}

\begin{table}[h]
\centering
\caption{Protocol layers, required inputs, and main outputs.}
\label{tab:layer_summary}
\small
\renewcommand{\arraystretch}{1.2}
\begin{tabular}{p{0.08\linewidth} p{0.31\linewidth} p{0.48\linewidth}}
\hline
\textbf{Layer} & \textbf{Inputs} & \textbf{Main outputs} \\
\hline
1 &
$z$, $p$ &
Linear and nonlinear probe $R^2$; disentanglement matrix;
random-Gaussian null baseline. \\

2 &
Flattened $X$,$z$,$p$ &
PCA baseline probe; per-parameter advantage
$A_j = R^2_{\mathrm{model},j} - R^2_{\mathrm{PCA},j}$. \\

3 &
Internal activations, $p$, $z$ &
Layer-wise physical decodability; 2D activation projections;
trajectory-level probe diagnostics. \\

4 &
Forward Network, $z$, $p$ &
Latent traversals; directional consistency $\rho_{k,j}$;
polynomial degree $D^\star$; normalized nudge effect
$\Delta \tilde{p}_j$. \\

5 &
Generated samples, thresholds &
Physical violation rate
$\mathrm{PVR}_{\mathrm{generated}}$ and shift
$\Delta_s =
\mathrm{PVR}_{\mathrm{generated}} -
\mathrm{PVR}_{\mathrm{train}}$. \\
\hline
\end{tabular}
\end{table}

\clearpage

\end{document}